\documentclass{article}

\usepackage{microtype}
\usepackage{graphicx}
\usepackage{subfigure}
\usepackage{booktabs} % for professional tables
\usepackage{amssymb,amsmath}
\usepackage{multirow}
\usepackage{caption}
\usepackage{textcomp}
\usepackage{enumitem}

\usepackage[table,xcdraw]{xcolor}
\definecolor{commentgreen}{RGB}{2,112,10}
\definecolor{eminence}{RGB}{108,48,130}
\definecolor{weborange}{RGB}{255,165,0}
\definecolor{frenchplum}{RGB}{129,20,83}

\usepackage{listings}
\PassOptionsToPackage{hyphens}{url}
\usepackage{hyperref}

\usepackage[accepted]{mlsys2022}

\mlsystitlerunning{Open-source FPGA-ML codesign for the MLPerf{\texttrademark} Tiny Benchmark}

\begin{document}

\twocolumn[

\mlsystitle{Open-source FPGA-ML codesign for the MLPerf{\texttrademark} Tiny Benchmark}

% \mlsyssetsymbol{equal}{*}

\begin{mlsysauthorlist}
\mlsysauthor{Hendrik Borras}{hu}
\mlsysauthor{Giuseppe Di Guglielmo}{columbia}
\mlsysauthor{Javier Duarte}{ucsd}
\mlsysauthor{Nicol\`{o} Ghielmetti}{cern}
\mlsysauthor{Ben Hawks}{fnal}
\mlsysauthor{Scott Hauck}{uw}
\mlsysauthor{Shih-Chieh Hsu}{uw}
\mlsysauthor{Ryan Kastner}{ucsd}
\mlsysauthor{Jason Liang}{ucsd}
\mlsysauthor{Andres Meza}{ucsd}
\mlsysauthor{Jules Muhizi}{fnal,harvard}
\mlsysauthor{Tai Nguyen}{ucsd}
\mlsysauthor{Rushil Roy}{ucsd}
\mlsysauthor{Nhan Tran}{fnal}
\mlsysauthor{Yaman Umuroglu}{amd}
\mlsysauthor{Olivia Weng}{ucsd}
\mlsysauthor{Aidan Yokuda}{uw}
\mlsysauthor{Michaela Blott}{amd}
\end{mlsysauthorlist}

\mlsysaffiliation{hu}{Heidelberg University, Heidelberg, Germany}
\mlsysaffiliation{columbia}{Columbia University, New York, NY, USA}
\mlsysaffiliation{harvard}{Harvard University, Cambridge, MA, USA}
\mlsysaffiliation{fnal}{Fermi National Accelerator Laboratory, Batavia, IL, USA}
\mlsysaffiliation{ucsd}{University of California San Diego, La Jolla, CA, USA}
\mlsysaffiliation{uw}{University of Washington, Seattle, WA, USA}
\mlsysaffiliation{amd}{AMD Adaptive and Embedded Computing Group (AECG) Labs, Dublin, Ireland}
\mlsysaffiliation{cern}{European Organization for Nuclear Research (CERN), Geneva, Switzerland}

\mlsyscorrespondingauthor{Javier Duarte}{jduarte@ucsd.edu}

\mlsyskeywords{Machine Learning, Science, Real-Time, Benchmark}

\vskip 0.3in

\begin{abstract}
We present our development experience and recent results for the MLPerf{\texttrademark} Tiny Inference Benchmark on field-programmable gate array (FPGA) platforms.  
We use the open-source hls4ml and FINN workflows, which aim to democratize AI-hardware codesign of optimized neural networks on FPGAs.
We present the design and implementation process for the keyword spotting, anomaly detection, and image classification benchmark tasks.
The resulting hardware implementations are quantized, configurable, spatial dataflow architectures tailored for speed and efficiency and introduce new generic optimizations and common workflows developed as a part of this work.
The full workflow is presented from quantization-aware training to FPGA implementation. 
The solutions are deployed on system-on-chip (Pynq-Z2) and pure FPGA (Arty A7-100T) platforms.  
The resulting submissions achieve latencies as low as 20\,$\mu$s and energy consumption as low as 30\,$\mu$J per inference.
We demonstrate how emerging ML benchmarks on heterogeneous hardware platforms can catalyze collaboration and the development of new techniques and more accessible tools.
\end{abstract}
]
\printAffiliationsAndNotice{} % otherwise use the standard text.

\section{Introduction}
\label{sec:intro}
Efficient implementations of machine learning (ML) algorithms in dedicated hardware devices at the \emph{edge}, or near sensor, offer multiple advantages.
Edge processing and data compression can greatly reduce downstream data rates and the energy required for data movement.
Furthermore, real-time data processing and interpretation can accelerate decision making, hypothesis testing, and enable just-in-time interventions.
These edge ML tasks can have a significant impact on a broad range of applications from \emph{internet of things} (IoT) to Industry 4.0~\cite{KagermannWahlsterHelbig2013en} and new experimental methods for scientific discovery~\cite{Deiana:2021niw}.

To enable broader adoption of these technologies, we present our solutions for the open division of the MLPerf\texttrademark~Tiny Inference Benchmark v0.7.
MLPerf Tiny has two divisions for submitting results: a stricter closed division and a more flexible open division, which allows submitters to alter ML model implementations and training workflows.
%For the closed division, submitters must use the same models, datasets, and quality targets as the reference implementation. 
%The open division allows submitters to change the model, training scripts, and dataset, while maintaining the same quality target.
We participated in the open division to demonstrate the advantages of hardware-AI codesign.

The hls4ml~\cite{Duarte:2018ite,Fahim:2021cic} and FINN~\cite{finn,blott2018finn,blott2018finnr} teams aim to democratize low-power, \emph{tiny}~\cite{tiny,tinymlbench}, accelerated ML by releasing accessible tools for the codesign of optimized neural networks on field-programmable gate arrays (FPGAs).
The hls4ml workflow originates from the Fast Machine Learning for Science community, which focuses on developing tools for scientific applications.  
FINN is an open-source project from AMD that enables the exploration of efficient ML acceleration on FPGAs.
These jointly developed solutions are the product of an ongoing collaboration between the FINN and hls4ml developers with the goal of making FPGA-accelerated tiny ML broadly available.

\begin{figure}[tbh!]
    \centering
    \includegraphics[width=0.99\columnwidth]{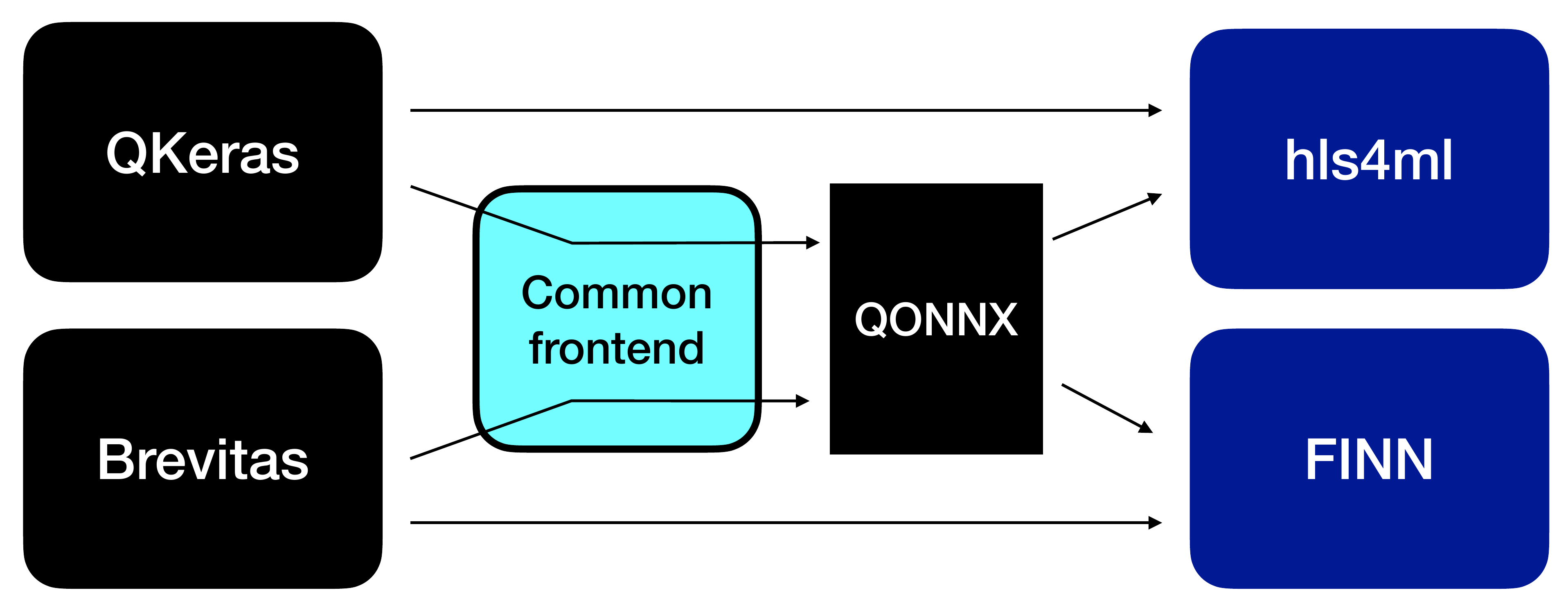}
    \caption{Common hls4ml-FINN FPGA codesign workflow based on QONNX.}
    \label{fig:flow}
\end{figure}

There are a number of unique features of the hls4ml and FINN workflows.  
Solutions support extreme flexibility in data type precision.
In fact, each solution from the team uses a different precision, from 1- to 12-bit operations.
The resulting hardware implementations are configurable, spatial, dataflow architectures that are tailored for speed and efficiency.  
The code, from end-to-end, is open-source and freely available including tools for design space exploration and the final implementations.
The workflow includes quantization-aware training (QAT) in QKeras~\cite{Coelho:2020zfu,qkeras} and Brevitas~\cite{brevitas}, hyperparameter optimization using Determined AI~\cite{det-ai} and KerasTuner~\cite{omalley2019kerastuner}, and FPGA implementation with hls4ml and FINN including Python APIs for inspection, validation, and deployment.  
Furthermore, one goal in these solutions is to develop a more unified workflow for quantized neural networks that is built on a common interchange format, called QONNX~\cite{qonnx_paper,qonnx}.  
The envisioned workflow is depicted in Fig.~\ref{fig:flow}.

The team consists of researchers collaborating across industry and academia.
The submissions are available on the TUL Pynq-Z2 platform with a Zynq-7020 system-on-chip (SoC) and Digilent Arty A7-100T platform with an Artix-7 100T FPGA.
The latter is the first submission on an FPGA-only platform.
Open division submissions are provided for the keyword spotting, image classification, and anomaly detection MLPerf Tiny benchmarks.
The resulting submissions on FPGA hardware were optimized for performance and speed with latencies as low as 20\,$\mu$s.  
The paper is structured as follows. 
In Section~\ref{sec:benchmarks}, we briefly describe the benchmark tasks. 
Section~\ref{sec:modeldev} presents the model optimization for performance, latency, and resources. The integration of those models into hardware is detailed in Section~\ref{sec:integration}. 
We summarize in Section~\ref{sec:summary}.

In this work, in the spirit of the MLPerf process, we do not provide detailed comparisons of our solutions with solutions on other hardware platforms.
However, the other submitted solutions for this benchmark can be found on the official MLPerf Tiny results page\footnote{ \url{https://mlcommons.org/en/inference-tiny-07}}.  

\subsection{Previous Work}
This work is built from a number of previous studies by the hls4ml~\cite{Duarte:2018ite,Summers:2020xiy,DiGuglielmo:2020eqx,Coelho:2020zfu,Aarrestad:2021zos,Iiyama:2020wap,Elabd:2021lgo} and FINN teams~\cite{finn,blott2018finn,blott2018finnr}.  
Other open-source efforts have explored ML-FPGA codesign.
Surveys of existing toolflows can be found in \citet{2018arXiv180305900V,10.1145/3289185,Shawahna_2019,abdelouahab2018accelerating}.
These workflows include fpgaConvNet~\cite{venieris2017fpgaconvnet,venieris2017fpga,venieris2017fpl,venieris2016fccm}, 
FP-DNN~\cite{fpdnn}, DNNWeaver~\cite{dnnweaver:micro16}, 
Caffeine~\cite{caffeinatedFPGAs}, 
Snowflake~\cite{snowflake}, Vitis AI~\cite{vitisai}, FixyNN~\cite{whatmough2019fixynn,deepfreeze}, and others~\cite{7459526,majumder2019flexible,hacene2018quantized,chang2020mix}.
The hls4ml and FINN workflows are unique with respect to other ML-to-FPGA workflows in two primary ways: the extreme configurability for low and arbitrary bit-precision and optimizing throughput using spatial dataflow architectures that resembles the flow of data through the NN on chip.  

\section{Benchmark Tasks and Models}
\label{sec:benchmarks}
The MLPerf Tiny benchmark consists of four tasks and reference implementations related to image classification (IC), anomaly detection (AD), keyword spotting (KWS), and visual wake words (VWWs).
We submitted solutions for the first three, which we describe in this section.
The full set of tasks is described in greater detail in~\citet{banbury2021mlperf}.

\subsection{Image classification}
Image classification is an important task for many autonomous and low-power embedded systems.
%New hardware platforms, algorithms and development tools can be used for these low-cost, low-latency, compact vision systems.
CIFAR-10~\cite{cifar10} is a labeled subset of the 80 Million Tiny Images dataset~\cite{torralba200880}. 
The low resolution of the images make CIFAR-10 suitable for training tiny image classification models. 
It consists of 60,000 $32\times32\times3$ RGB images, with 6,000 images per class. 
The 10 different classes represent airplanes, cars, birds, cats, deers, dogs, frogs, horses, ships and trucks. 
The dataset is divided into 50,000 training images and 10,000 testing images.

The reference IC model for the closed division is a customized version of ResNetV1~\cite{resnet,banbury2021mlperf} which takes as input $32\times32\times3$ images and outputs a probability vector of size 10. 
This customized model is composed of three residual residual stacks rather than four. 
Each stack consists of three convolutional layers.
Moreover, the first convolutional layer is not followed by the pooling layer due to the low resolution of the input data. 
The number of convolutional filters and strides are also lower compared to the official ResNet.

A subset of 200 images from the CIFAR-10 test set are selected to evaluate the performances of the IC reference implementation. 
For the v0.5 benchmark a class-unbalanced subset was chosen, whereas for the v0.7 benchmark the subset was updated to maintain class balance.
The reference model achieves 87.0\% accuracy across the 200 testing images of the v0.7 benchmark.

\subsection{Anomaly detection}
Anomaly detection is a task that requires separating normal and anomalous signals in various data formats. 
For this benchmark as defined by MLPerf Tiny, an unsupervised approach is developed to train the neural network to closely match industrial use-cases where normal behavior may be well defined because it is less feasible to collect every possible anomalous signal and train a binary classifier in a supervised learning approach.

The unsupervised AD model is trained on the DCASE 2020 Challenge Task 2 dataset which employs the  ToyADMOS~\cite{koizumi2019toyadmos} ToyCar dataset. 
The dataset is comprised of 10\,s WAV files. 
The full set is split into 7,000 normal audio files for training and 2,459 for testing.

Before training on the audio files, we preprocess them into mel spectrograms of 128 bands describing each 32\,ms interval. 
The model is then trained on a sliding window of five frames of the spectrogram yielding an input size of 640. 
We use an autoencoder NN structure that attempts to recreate the input. 
We calculate the mean-squared error (MSE) between the input and output and average it over each of the windows (196$\times$). 
We use a smaller version of the MLPerf Tiny AD autoencoder network that has 128 inputs with an encoder and decoder comprised of two quantized 72-unit fully-connected (FC) layers with batch normalization (BN) and ReLU activation.
An FC layer is also used as the output layer. 
To evaluate the model performance, we average the MSE over each the windows in the audio sample to compute an anomaly score. 
To set the threshold between normal and anomalous sounds, we use the receiver operating characteristic (ROC) curve and the corresponding area under the curve (AUC) as the quality metric.

\subsection{Keyword spotting}

Over the last decade, keyword spotting has become increasingly prevalent, especially in modern voice assistants. 
Running a full speech recognition system only to detect an activation word is often impractical because of power implications and privacy concerns.
Instead, modern devices only listen for an activation word, a specific keyword. 
Since recognizing a limited set of words is significantly simpler than full speech recognition, the keyword spotting system can be run locally on a given device and with low power impact.
Keyword spotting is also interesting for general robot control, by setting a vocabulary with words such as ``start,'' ``stop,'' ``louder,'' and ``quieter.''

For the MLPerf Tiny benchmark the KWS task is based on the Google speech commands dataset V2~\cite{speechcommandsv2}.
The dataset consist mainly of 1\,s audio files, each containing one spoken word.
In total 105\,829 data samples are available, recorded by different speakers. 
The data samples are partitioned into training, validation, and test sets, such that a given speaker only appears in one of the sets. 
Additionally, longer files with background noises are included.
Overall the dataset contains 35 classes, each representing the utterance of one word. 
The dataset is however more often used in its twelve-class variant, where ten fixed classes are used and the additional 25 classes are grouped into the unknown class, while also adding a new class called silence, comprised of samples from the included background noises.
For MLPerf Tiny, this twelve-class variant is used.
Along with sample code for setting up the pre-processing for the dataset MLPerf Tiny also provides a reference model for the KWS task, which is a depthwise separable convolutional neural network (DS-CNN)  from~\citet{helloEdge}. 
The DS-CNN model is relatively compact and optimized for low-power microcontrollers.
For on-device testing, 1,000 samples from the Google speech commands test test are selected for the benchmark. 
Over this subset, the reference model achieves an accuracy of 92.2\%.

\section{Model development and codesign}
\label{sec:modeldev}
An overview of the models developed for the submission are presented in Table~\ref{tab:summary}.  
Two models for the IC task one each for the AD and KWS tasks were submitted.
Below, we will describe in detail the model architecture and optimizations, both for training and implementation, that were performed for each of the models for the two FPGA hardware platforms considered.  
To optimize performance, all of the FPGA neural network do not use off-chip memory. 
However, to measure the performance of the models using the MLPerf Tiny benchmarking suite, the ARM processing system uses off-chip memory which would not necessarily be required in a standalone design.
Each model is developed through hardware-software codesign to search for Pareto-optimal solutions in model accuracy and resource usage by tuning a number of design machine learning and hardware architecture parameters.  
The solutions are not configurable at runtime for optimized performance. 

\begin{table}[htpb]
\centering
\resizebox{\columnwidth}{!}{
\begin{tabular}{l|l|r|r|r}
Benchmark & Flow & Prec. [bits] & Params. & Accuracy \\
\hline
IC & hls4ml & 8--12 & 58\,115 & 83.5\% \\
IC & FINN & 1 & 1\,542\,848  & 84.5\% \\
AD & hls4ml & 6--12 & 22\,285 & 0.83 AUC \\
KWS & FINN & 3 & 259\,584 & 82.5\%
\end{tabular}
}
\caption{Summary of models submitted for the v0.7 benchmark including benchmark task, tool flow used, precision of model, number of parameters, and performance---by default this is accuracy unless denoted as AUC}
\label{tab:summary}
\end{table}

\subsection{Optimization for IC with hls4ml}

To find models that simultaneously accomplished the goals of high accuracy, low latency, low resource usage (such that they can be accommodated on the chosen FPGA platforms), and low power utilization, a sequence of neural architecture search (NAS), QAT with QKeras, configuration and implementation with hls4ml, and model- and hardware-centric optimizations were performed.

\subsubsection{Bayesian Optimization}
For the NAS, the MLPerf Tiny benchmark reference ResNet-8 model was chosen as a starting point. 
The model was generalized in several ways to allow a restricted NAS.
In particular the tunable hyperparameters include the total number of stacks, the number of filters, filter size, and strides in each convolutional layer, whether average pooling is applied before the final dense layer, and whether whether skip connections are enabled. 

\begin{figure}[htbp]
    \centering
    \includegraphics[width=0.99\columnwidth]{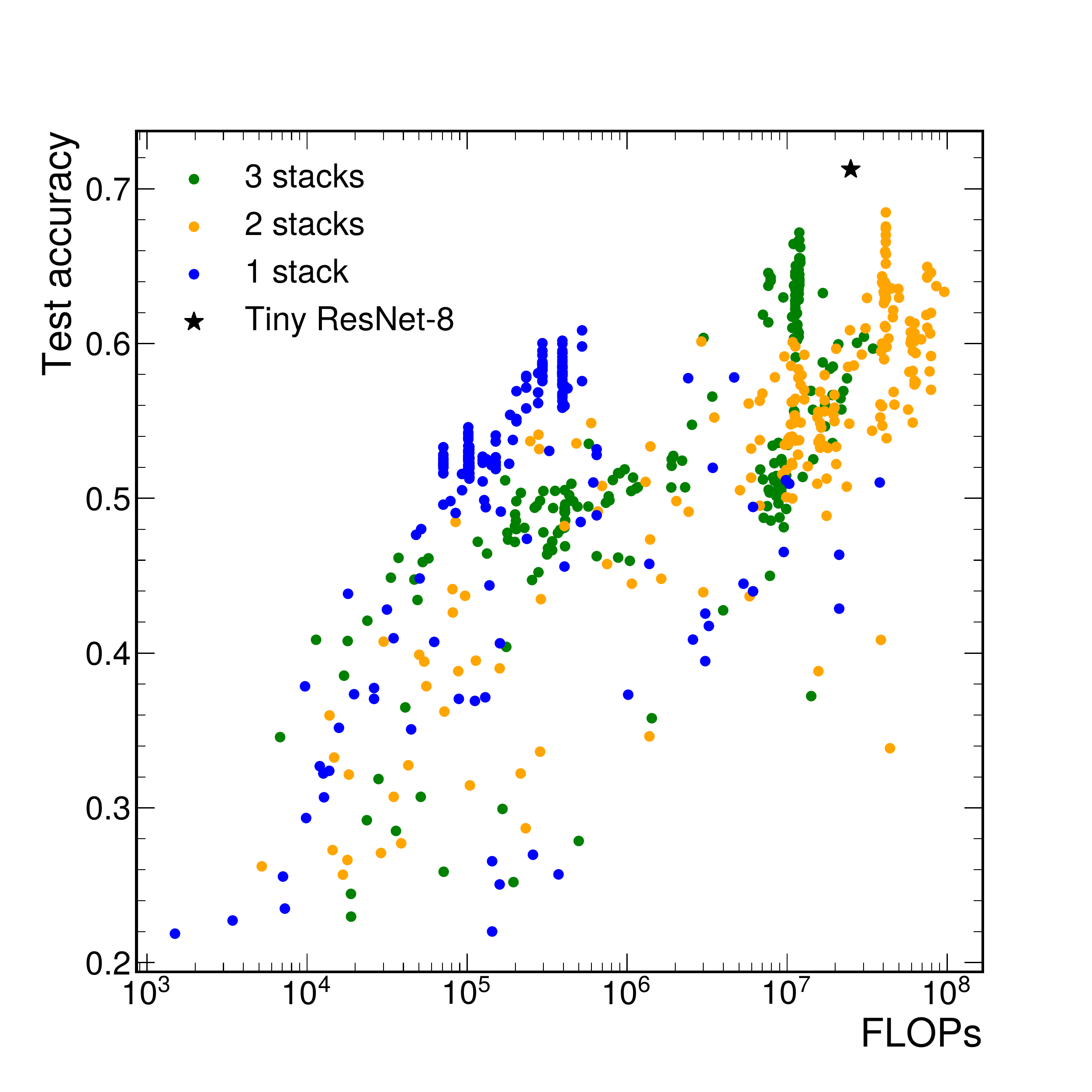}
    \caption{Results of the BO scans for 1-, 2-, and 3-stack models.
    The star represents the test accuracy achieved after 10 epochs with tiny ResNet-8 model }
    \label{fig:bo}
\end{figure}

We perform several Bayesian optimization (BO) scans using KerasTuner. 
We consider 1-, 2-, and 3-stack models in separate scans of 100 models each.
For each scan, we consider 2, 4, 8, or 16 filters, filter sizes of 1, 2, or 3, and particular strides to allow for valid skip connections.
We use a batch size of 32 and train each model for 10 epochs.
During training, the input data is normalized by dividing by 256.
For each model, we compute the best test accuracy and the number of floating point operations (FLOPs)~\cite{kerop}.
The results of these BO scans are shown in Fig.~\ref{fig:bo}.
We generally find that 1-stack models generally provide a good balance of a smaller number of FLOPs, while maintaining high accuracy.
The number of filters has the biggest impact on the accuracy and FLOPs of the models. 
Larger stride lengths and smaller filter sizes can also reduce FLOPs at the cost of some accuracy.
Neither average nor max pooling gave significant improvement.

BO allowed us to narrow down our choices to a very few models by revealing the most important hyperparamter values.
From the results of our scans, we found a 1-stack model (3 convolutional layers  with 32, 32, and 32 filters, kernel sizes of 3, 3, and 3, and strides of 4, 4, and 1, respectively, and an FC layer with 2048 units) used for the v0.5 submission that achieves a test accuracy of 75.0\% for 2.5 MFLOPs and 12.8 MFLOPs, respectively.
We also found a 2-stack model with no skip connections (5 convolutional layers, with 32, 4, 32, 32, and 4 filters, kernel sizes of 1, 4, 4, 4, and 4, and strides of 1, 1, 1, 4, 1, respectively, and an FC layer with 2048 units) used for the v0.7 submission that achieves a test accuracy 83.5\% for 12.8 MFLOPs.
Compared to the reference model, which corresponds to an accuracy of 87.0\% for 25.0 MFLOPs, these optimized models represent a substantial reduction of FLOPs for a minor reduction in accuracy. 
After settling the model architecture, we performed QAT with QKeras~\cite{qkeras,Coelho:2020zfu} using a fixed-point precision of 8 total and 2 integer bits.
Because the benchmark only measures the top-1 accuracy of the model, it is only necessary to return the class predicted to be highest probability.
Since softmax layer is monotonic in the input logits, it is not necessary to find the top-predicted class (i.e. a simple $\max$ applied to the logits is sufficient), and thus it is  removed for inference.
%The optimized model architectures are shown in Fig.~\ref{fig:RN_diagrams}, compared to the baseline reference model for the closed division.

%\begin{figure}
%    \centering
%    \includegraphics[width=.99\columnwidth]{Figs/RN_diagrams.pdf}
%    \caption{IC models for (a) baseline reference, (b) v0.5 submission, and (c) v0.7 submission -- \textcolor{red}{can we say this in words -- we don't do this for other models and it's impossible to read anyway?}} \label{fig:RN_diagrams}
%\end{figure}

\subsubsection{FIFO Buffer Depth Optimization}
\label{sec:ic:hls4ml:fifo}
By exploiting the dataflow preprocessor directive (or \emph{pragma}) of Vivado HLS, each layer composing a neural network is connected with the rest of the model through first-in first-out (FIFO) buffers. 
The implementation of the FIFO buffers contribute to the overall resource utilization of the design, impacting in particular the block random access memories (BRAMs) or look-up table (LUT) utilization. 
Because the neural networks can have complex architectures generally, is hard to know a priori the correct depth of each FIFO buffer. 
In order to reduce the impact on the resources used for FIFO buffer implementation, an optimization has been developed which aims to correctly size the depth of the FIFO buffers by analyzing the data produced by the register-transfer level (RTL) simulation. 
We implemented this FIFO buffer resizing within the hls4ml framework as an optimization pass. 
%Through the analysis of a value change dump file generated by the RTL simulation, it is possible to estimate the maximum occupation of each FIFO buffer, performing the simulation on large FIFO buffers (by default the size is 10\,k elements but it can be modified). 
Through RTL simulation with large FIFO buffers, we estimate the maximum occupation of each FIFO.
%, performing the simulation on large FIFO buffers (by default the size is 10\,k elements but it can be modified). 
Once the maximum depth is determined, the optimization pass sets the FIFO buffer depth to that value plus 1. %, for each FIFO buffer. 
%proved useful, giving the possibility of deploying models that until now could not be implemented on our target FPGAs due to lack of resources. 
In Table~\ref{tab:opt}, the FPGA resource usage for the hls4ml IC model is shown with and without the optimization.
This optimization significantly reduces the FPGA resources enabling the deployment of larger models.

For FINN an equivalent optimization exists, which was applied to all FINN-based models in this submission.
Fundamentally the optimization executes very similar steps to the optimization in hls4ml, running PyVerilator~\cite{pyverilator} on the full design, to perform an estimation for the optimal FIFO buffer depths between the layers of a given neural network design. 
The found FIFO buffer depths are then saved in the internal ONNX representation and are applied at a later step. 
Even though the simulation of the whole model in an RTL simulation is time consuming, this approach has proven useful for many FINN models, such that it is now part of the default compiler flow in FINN.

Table ~\ref{tab:fifo_summary} shows a summary of the FIFO buffer sizes set with this optimization for both hls4ml and FINN.

\begin{table}[htpb]
\centering
\resizebox{\columnwidth}{!}{
\begin{tabular}{l|l|l|l}
Benchmark & Flow & FIFO optimization & FIFO size \\
\hline
IC & hls4ml & enabled & 1--1066 \\
IC & FINN & enabled & 2--512 \\
AD & hls4ml & disabled & 1 \\
KWS & FINN & enabled & 32--64
\end{tabular}
}
\caption{Summary of FIFO buffer sizes for models submitted for the v0.7 benchmark. 
For the hls4ml FIFO optimization the FIFO buffer sizes can take an arbitrary integer values, while for FINN they can only be powers of two. 
No FIFO optimization was performed for the AD model.}
\label{tab:fifo_summary}
\end{table}

\subsubsection{ReLU Layer Merging}
As mentioned in the previous section, each dataflow stage consists of a neural network layer, which  are linked together by FIFOs that cost BRAMs, LUTs, and flip flops (FFs).
By default in hls4ml, each rectified linear unit (ReLU) layer is implemented as its own dataflow stage. 
Because each additional dataflow stage costs extra logic and FIFOs, we reduce the resource utilization by merging the ReLU activation function into the layer preceding it. 
%By merging these two dataflow stages into one, we reduce the overall number of dataflow stages and FIFOs by the number of ReLU functions, decreasing the overall resource utilization. 
Although the layers with the newly merged ReLU functionality use more logic than before, there is still a net decrease in resources.
%We merge the ReLU layers into their preceding convolutional layers, leading to a decrease in BRAMs, FFs, and LUTs. 
Table~\ref{tab:opt} shows the resulting resource utilization reductions.
%The final implemented design on the Pynq-Z2 device is shown in Fig.~\ref{fig:rn07_impl}

\begin{table}[htpb]
    \centering
\resizebox{\columnwidth}{!}{
    \begin{tabular}{l|rr|rr|rr}
& \multicolumn{2}{c|}{BRAM [18\,kb]}	&	\multicolumn{2}{c|}{FF}		&	\multicolumn{2}{c}{LUT}	\\
Available	& \multicolumn{2}{c|}{280}	&	\multicolumn{2}{c|}{106\,400} &		\multicolumn{2}{c}{53\,200}	\\\hline
Without opt. &	477 &	170.4\%	& 79\,177 &	74.4\%	& 66\,838 &	125.6\% \\
With FIFO opt. & 	278	& 99.3\% &	72\,686 &	68.3\% &	58\,515 &	110.0\%\\
With ReLU opt. &	345 &	123.2\%	& 72\,921 &	68.5\% &	55\,292 &	103.9\%\\
With all opt. &	146 &	52.1\% & 66\,430 &	62.4\% &	46\,969 &	88.3\%
    \end{tabular}}
    \caption{Resource estimates from Vivado HLS for the IC model with hls4ml for the v0.7 MLPerf Tiny submission }
    \label{tab:opt}
\end{table}

%\begin{figure}
%    \centering
%    \includegraphics[width=0.99\columnwidth]{Figs/RN07.png}
%    \caption{Implemented design of the IC model generated with hls4ml on the PYNQ-Z2 device.}
%    \label{fig:rn07_impl}
%\end{figure}

\subsection{Optimization for IC with FINN}

The model submitted for the IC task with FINN is called CNV-W1A1 from~\citet{finn}.
The model architecture takes inspirations from BinaryNet~\cite{binaryNet} and VGG-16~\cite{VGG}, consisting of first multiple convolutional blocks and then fully connected layers at the end. 
The whole architecture can be described as follows:
\begin{itemize}[noitemsep,topsep=0pt]
    \item Three convolutional blocks, consisting of two $3\times 3$ convolutions and one $2\times 2$ max pooling layer at the end. 
    The convolutions in each of these blocks have the following number of channels respectively: 64, 128, 256.
    \item The network then continues with two fully connected layers with 512 neurons and one output layer with 10 neurons.
    \item Finally a top-k layer is inserted to calculate the classification result in hardware.
\end{itemize}
Since the original release of the FINN paper the framework has been extended to support arbitrary bit widths, meaning that weight and activations with more than one bit can also be synthesized.
However, bit widths below eight bit are generally recommended for FINN, due to how the underlying activation implementation scales with bit width.
As such the CNV model also exists in variants with two bit weights and activations. 
For the MLPerf Tiny submission, the binary version of the model is used.
Here, the weights and activations are quantized to a bipolar representation, with the notable exception being the input layer, which processes the input images as 8-bit data. Consequently the activation function associated with the input layer performs an eight bit calculation, while all other layers of the network work with a binary representation of the weights and activations.

\subsubsection{ASHA for IC with FINN}

We used the adaptive ASHA algorithm~\cite{aASHA} from Determined AI to search for a more efficient or accurate model.
The starting point for the scan was the CNV-W1A1 model. 
The hyperparameters that were varied were the number of convolutional filters (from 32 to 512), whether or not to pool after convoluational layers, strides (from 1 to 4), kernel sizes (from 1 to 4), pooling size (2 or 4), number of neurons in fully connected layers (from 16 to 512), and activation and weight bit widths (1 or 2),
The adaptive scan allocates a set of resources to scan varied hyperparameter configurations, throwing out the worse half based on the specified validation metric and repeating until only one optimal configuration remains. 
A batch size of 50 was used and each model was trained for up to 100 epochs, although the adaptive ASHA algorithm may terminate training earlier.

\begin{figure}
    \centering
    \includegraphics[width=0.99\columnwidth]{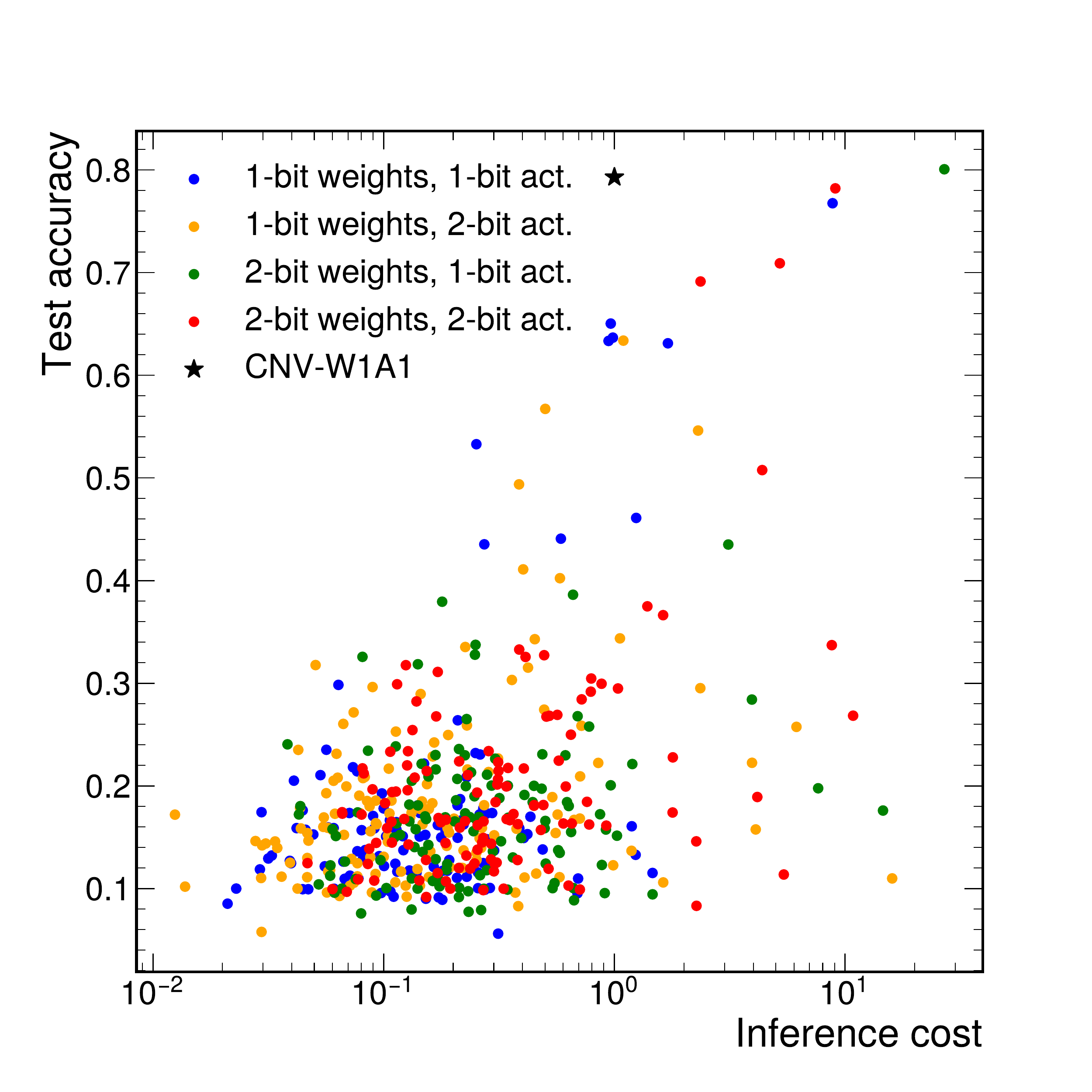}
    \caption{Results of ASHA scan in terms of accuracy as a function of the inference cost hardware metric. 
    The CNV-W1A1 used in the submission is shown for reference (inference cost of 1), with its test accuracy after 100 epochs of training}
    \label{fig:asha}
\end{figure}

For each model, several inference cost metrics are computed including total bit operations (BOPs)~\cite{bops,Hawks:2021ruw} and the total number of bits needed to store the weights in memory (WM).
BOPs count the number of multiply-accumulate operations in a neural network multiplied with the bit width at which this operation is performed. 
For a single convolutional layer with $b_\mathrm{w}$-bit weights and $b_\mathrm{a}$-bit activations containing $n$ input channels, $m$ output channels, and $k \times k$ filters,
\begin{equation}
	\text{BOPs} \approx mnk^2( b_\mathrm{a} b_\mathrm{w} + b_\mathrm{a}+b_\mathrm{w} + \log_2 {nk^2})\,.
\label{eq:bops}
\end{equation}
For computing BOPs for fully connected layers, we set $k=1$ in Eq.~\ref{eq:bops}.  
This metric functions as a preliminary estimate for the FPGA resource usage of the network implemented with FINN and thus allows for a first comparison between networks before any synthesis takes place.
These two metrics, BOPs and WM, were inspired by the ITU AI for Good Challenge: Lightning-Fast Modulation Classification with Hardware-Efficient Neural Networks~\cite{itu}.
A summary inference cost metric also inspired from that competition is defined as
\begin{equation}
 C = \frac{1}{2}\left(\frac{\text{BOPs}}{\text{BOPs}_\text{CNV-W1A1}} + \frac{\text{WM}}{\text{WM}_\text{CNV-W1A1}}\right)\,,
\end{equation}
where the CNV-W1A1 model is taken as reference for comparison.
Figure~\ref{fig:asha} shows the result of the scan in terms of accuracy as a function of the inference cost.
Based on our results, the CNV-W1A1 model performs near optimally.

\subsection{Optimization for AD with hls4ml}
When implementing the reference AD multilayer perceptron (MLP) model on the Pynq-Z2 and Arty A7-100T platforms, the limiting resource was LUTs. 
The reference floating point implementation was too large to synthesize, therefore, we optimized the architecture using quantization and model compression techniques with minimal AUC performance reduction.
As a part of this process, we implemented a generic optimization to fold the batch normalization (BN) into the FC layer and reduced the the depth and width of the autoencoder network.

%A major challenge of this benchmark task is running a low latency and low power solution.
%While there are algorithmic and hardware platform level choices that can be made to decrease the latency, we target the Pynq-Z2 and Arty A7-100T FPGA, which come with a constrained number of look-up table (LUT) units. 
%In order to be within the LUT constraints, we performed a optimization attempting the reduce the footprint size of the model via a combination of quantization, folding batch normalization into the FC layer and reducing the the depth and width of the autoencoder network.

\subsubsection{QDenseBatchnorm Layer}
To remove the LUT utilization from the BN computation, we developed a new quantized FC layer in QKeras~\cite{qdensebatchnorm} where during the forward pass, we compute the matrix multiplication of the FC layer as well as BN in the same pass and then save a kernel that ``folds'' the FC kernel with the BN parameters into a folded kernel by transforming the dense kernel with the BN parameters as defined in equation~\ref{eq2}.
Furthermore, within the same pass, both the BN parameters as well as the folded kernel are updated. 
During the forward pass, we compute the FC layer outputs then pass the outputs into a BN pass in order to update the BN parameters. 
Once we have the new BN parameters, we then fold them into the FC layer parameters as,
\begin{align}
k_\mathrm{folded} &= v k_\mathrm{FC}\label{eq2}\\
b_\mathrm{folded} &= v (b_\mathrm{FC} - \mu) + \beta
\end{align}
where $v = \gamma\sqrt{\sigma^2 + \epsilon}$ with $\mu$ and $\sigma$ the moving mean and standard deviation and $\epsilon$ and $\beta$ the BN scale and shift parameters. $\gamma$ represents a learned scale factor in the original kernel. $k_\mathrm{FC}$ ($k_\mathrm{folded}$) represents the original (folded) kernel, and $b_\mathrm{FC}$ ($b_\mathrm{folded}$) is the original (folded) bias. 

\subsubsection{Reuse Factor, Input Size, and Layer Depth}

To further tune the LUT utilization for the model, we varied the parallelization of the algorithm via the \textit{reuse factor} (RF), or how many times each multiplier unit is used, which controls the parallelization of the algorithm. 
With the goal of a low latency solution, we optimized for the lowest RF factor.
We then performed a scan across the available RFs and synthesized onto the Pynq-Z2, while also tracking the overall resource utilization on the FPGA. 
After the scan, we found that the smallest RF deployable on the FPGA is 144. 
Table~\ref{tab:ad_opt} summarizes the optimizations from the reference model to the final submitted model.
By reducing the number of hidden layers and width of each layer from 9 to 5 and 128 to 72, respectively, we reduce the key bottleneck, the LUT count, to 161\,228. 
Combined with downsampling of the input from 640 to 128, our optimizations achieve a final 58.5\% utilization of the FPGA LUTs on the Pynq-Z2.

\begin{table}[htpb]
    \centering
\resizebox{\columnwidth}{!}{
    \begin{tabular}{l|r|rr|rr}
& AUC	&	\multicolumn{2}{c|}{FF}		&	\multicolumn{2}{c}{LUT}	\\
Available	& --	&	\multicolumn{2}{c|}{106\,400} &		\multicolumn{2}{c}{53\,200}	\\\hline
Reference & 87.1\% 	& -- &	--	& -- &	-- \\
With folding &	68.1\%	& 161\,228 &	151.5\%	& 221\,063 &	451.5\%\\
With downsampling  &	81.4\% & 55\,341 &	52.0\% &	35\,366 &	66.5\%\\
With all opt.	& 83.3\% &	44\,300 &	41.6\% &	31\,094 &	58.5\%\\
    \end{tabular}}
    \caption{Resource utilization from Vivado HLS logic synthesis on the Pynq-Z2 for the hls4ml AD model with various optimizations at 144 reuse factor.
    \label{tab:ad_opt}}
\end{table}

\subsection{Optimization for KWS with FINN}
% Script for creating this plot: https://gist.github.com/HenniOVP/bdd5dff585480aae709eb1e5bd608006
% and updated by Javier here: https://gist.github.com/jmduarte/cd71b3dabc1501da5d5e41248f99c70b
\begin{figure}[tbh!]
    \centering
    \includegraphics[width=0.99\columnwidth]{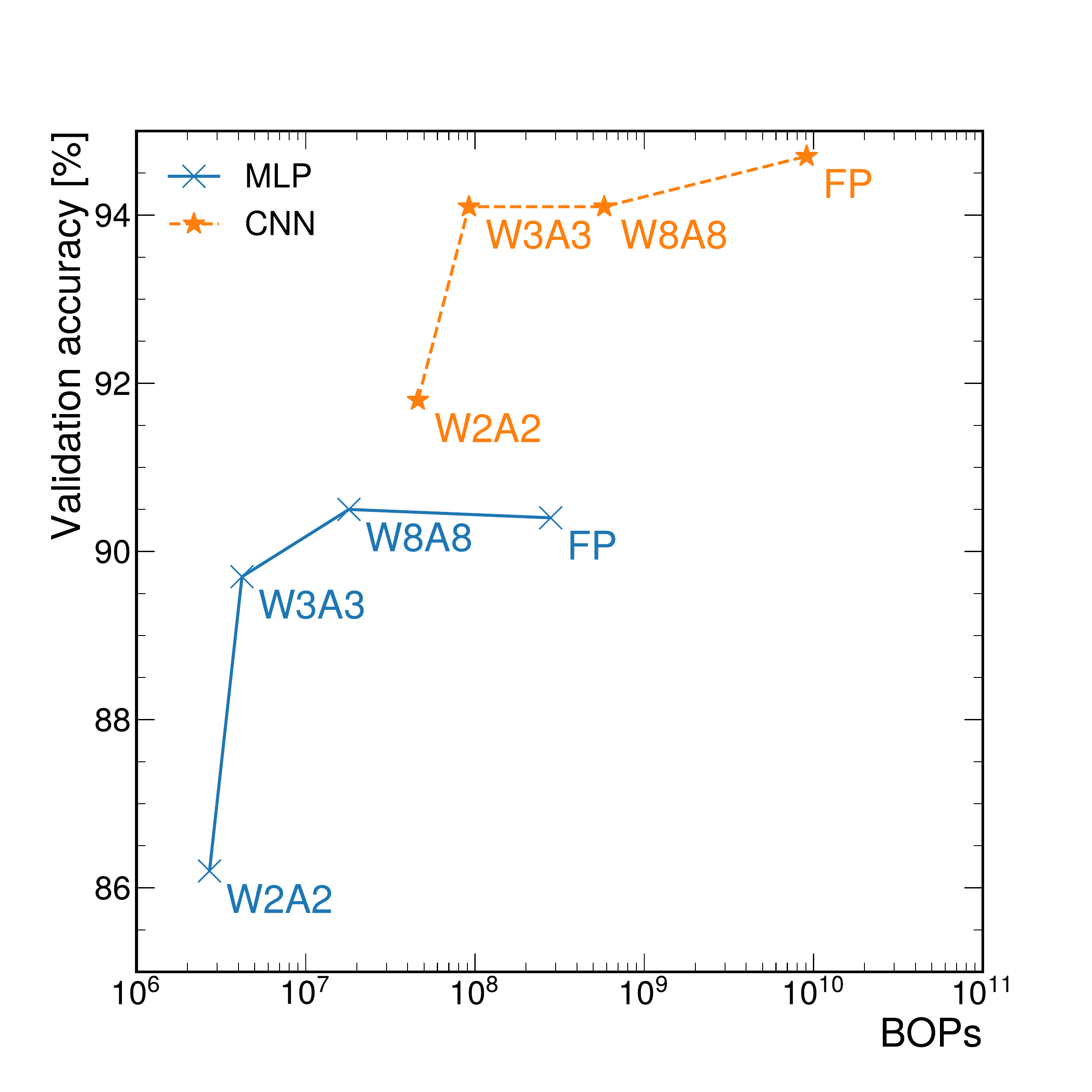}
    \caption{Quantization exploration for keyword spotting. 
    Each data point is annotated with its weight and activation bit width, in the following schema: W$n$A$m$, with $n$ the weight bit width and $m$ the activation bit width. 
    Additionally, floating point data points are annotated with FP. 
    The model accuracy on the validation dataset is shown on as a function of the network complexity quantified by BOPs.
    \label{fig:kws_quant}}
\end{figure}
The model submitted for keyword spotting and synthesized by FINN is directly inspired by~\citet{helloEdge}.
Initially, two architectures from this paper were explored, the MLP and the convolutional neural network (CNN).
However, given the more complex rectangular convolutions employed in the CNN, the MLP was chosen for its simplicity.
The architecture of the MLP consists of three fully connected (FC) layers, each with BN and ReLU activations. 
One final output layer with 10 neurons is also used. 
Similar to the image classification network synthesized with FINN an in-hardware top-k node was inserted at the end.

A weighted cross-entropy loss was employed during training. 
The re-weighting largely suppresses the unknown label in the dataset to combat the imbalance between classes, where for the 12 class version of the google speech commands V2 dataset, the unknown label is present about 17 times more often than any other label. 
The exact suppression setting for the unknown label was then found by running an adaptive ASHA hyperparameter search~\cite{aASHA}. 
Additionally, the training was managed using Determined AI~\cite{det-ai}, which allowed for easy integration of the adaptive ASHA algorithm into the general training flow.
For the feature extraction, mel-frequency cepstral coefficients (MFCC) were adapted and implemented as done for the closed division reference implementation.
The primary optimizations for the KWS submission included investigating the different quantization settings for activations and weights.
QAT was performed with Brevitas~\cite{brevitas}, which is an extension to the popular machine learning framework PyTorch~\cite{pytorch}.
The exploration process is shown in Fig.~\ref{fig:kws_quant}. 
Here, the model accuracy on the validation dataset is plotted against the network complexity quantified by BOPs.
To find the appropriate quantization for the network, a reference model was first trained at floating-point precision. 
After this, the bit widths of the weights and activations were successively lowered, until the network validation accuracy dropped significantly. 
This sudden decrease was found for both the CNN and MLP to be below three bits for weights and activations. 
Thus, 3-bit quantization was chosen for the submission.
However, notably the network input is 8 bits. %, meaning that the first layer operates on eight bit inputs, allowing for an overall higher network accuracy.

\subsection{Automatic Optimizations for FINN Models}
In addition to the optimizations applied to the IC and KWS models described above, FINN automatically applies multiple optimizations before a design is synthesized.

As a first step, FINN applies constant folding. 
Here, constant initialized tensors are propagated through the ONNX~\cite{onnx} computation graph, and nodes that return the constant values are precomputed at compile time.
This basic optimization can save significant compute overhead and makes most networks easier for the compiler to handle.

Afterwards, a graph transformation called streamlining is applied. 
This transformation is a direct application of~\citet{streamlining}. 
The operation folds layers, which are usually computed at floating-point precision into integer operations for uniformly quantized neural networks. 
For FINN, this primarily affects BN layers, which are folded into multi-threshold nodes, which can represent arbitrarily quantized activation functions. 
In addition to removing time-consuming floating-point operations, this optimization eliminates some runtime computations entirely.

Before layers are converted into generated HLS code, FINN will minimize the final accumulator datatypes for all threshold-based operations. 
In particular, FINN minimizes the memory footprint for all activation layers.
After the individual layers of a network have been synthesized into Vivado intellectual property (IP) blocks, FINN performs a FIFO buffer optimization to balance its dataflow pipeline. 
This method is conceptually similar to the recently implemented optimization pass in hls4ml (see Sec.~\ref{sec:ic:hls4ml:fifo}) and achieves similar performance improvements.

% \subsection{Models overview}
% \begin{itemize}
%     \item \textcolor{red}{Just an idea, please chime in!}
%     \item \textcolor{red}{A table here might be interesting with: dataset, accuracy, precision/s used, op and param count for all models to give a sense of scale to all these models.}
% \end{itemize}

\section{System integration}
\label{sec:integration}
\subsection{QONNX Interchange Format}
% (Hendrik)

Currently, all networks synthesized with hls4ml were trained with QKeras and all submissions synthesized with FINN were trained with Brevitas.
In the future, we plan to be able to synthesize models trained with either quantization aware training (QAT) library in both compiler frameworks.
The central component to this process is an interchange format called QONNX~\cite{qonnx_paper,qonnx}, which is an extension to the ONNX standard~\cite{onnx}. 
It introduces new quantization nodes for arbitrary uniform quantization, as is required by both frameworks.
Already now FINN and Brevitas both fully support the QONNX format, since version 0.7 of both frameworks, and the KWS submission already uses this format. 
QONNX should enable simpler and faster exchange of QAT models between different FPGA-ML flows like hls4ml and FINN. 
%For example, if progress is made for the training of a given model in any QAT tool, that the new model can then immediately be run in either of the synthesis frameworks and progress can be shared more easily.

\subsection{ML Accelerators}

% \begin{figure*}[!htb]
%     \centering
%     \begin{minipage}{.5\textwidth}
% \begin{lstlisting}
% void top_module_axi(input_axi_t in[N_IN], output_axi_t out[N_OUT]) {
%   #pragma HLS INTERFACE s_axilite port=return bundle=CTRL_BUS
%   #pragma HLS INTERFACE m_axi port=in offset=slave bundle=IN_BUS
%   #pragma HLS INTERFACE m_axi port=out offset=slave bundle=OUT_BUS
  
%   hls::stream<input_t> in_local("input_1");
%   #pragma HLS STREAM variable=in_local depth=N_IN_LOCAL
%   hls::stream<output_t> out_local("output_1");
%   #pragma HLS STREAM variable=out_local depth=N_OUT_LOCAL
 
%   for(unsigned i = 0; i < N_IN / input_t::size; ++i) {
%     input_t ctype;
%     #pragma HLS DATA_PACK variable=ctype
%     for(unsigned j = 0; j < input_t::size; j++) {
%       ap_ufixed<16,8> tmp = in[i * input_t::size + j];
%       ctype[j] = typename input_t::value_type(tmp >> 8);
%     }
%     in_local.write(ctype);
%   }
  
%   module(in_local, out_local);
  
%   for(unsigned i = 0; i < N_OUT / output_t::size; ++i) {
%     output_t ctype = out_local.read();
%     for(unsigned j = 0; j < output_t::size; j++) {
%       out[i * output_t::size + j] = output_axi_t(ctype[j]);
%     }
%   }
% }
% \end{lstlisting}
%     \end{minipage}%
%     \begin{minipage}{0.5\textwidth}
%         \centering
%         \includegraphics[width=0.5\columnwidth]{Figs/hls4ml_interface.pdf}\\
%         \vspace{5px}
%         \includegraphics[width=0.8\columnwidth]{Figs/top_module_axi.png}\\
%     \end{minipage}
% \caption{hls4ml: HLS code for the top level module, accelerator architecture, and accelerator interface}
% \label{fig:hls4ml_accelerator}
% \end{figure*}

\subsubsection{hls4ml and FINN for Dataflow Architectures}
%FINN (Hendrik) / hls4ml (Giuseppe) accelerator IPs}

%\begin{itemize}
%    \item \textcolor{red}{Describe the structure of a FINN and hls4ml accelerator, dataflow-based accelerators, data-movers, AXI interface, accelerator local buffers vs. off-chip-memory etc.}
%    \item \textcolor{red}{Merge and highlight the similarities between FINN and hls4ml.}
%\end{itemize}

\begin{figure}[t!]
    \centering
    \begin{minipage}{0.45\textwidth}
    \footnotesize
\begin{lstlisting}
void top_module_axi(input_axi_t in[N_IN], output_axi_t out[N_OUT]) 
{
  #pragma HLS INTERFACE s_axilite port=return bundle=CTRL_BUS
  #pragma HLS INTERFACE m_axi port=in offset=slave bundle=IN_BUS
  #pragma HLS INTERFACE m_axi port=out offset=slave bundle=OUT_BUS
  
  hls::stream<input_t> in_local("input_1");
  #pragma HLS STREAM variable=in_local depth=N_IN_LOCAL
  hls::stream<output_t> out_local("output_1");
  #pragma HLS STREAM variable=out_local depth=N_OUT_LOCAL
 
  for(unsigned i = 0; i < N_IN / input_t::size; ++i) {
    input_t ctype;
    #pragma HLS DATA_PACK variable=ctype
    for(unsigned j = 0; j < input_t::size; j++) {
      ap_ufixed<16,8> tmp = in[i * input_t::size + j];
      ctype[j] = typename input_t::value_type(tmp >> 8);
    }
    in_local.write(ctype);
  }
  
  module(in_local, out_local);
  
  for(unsigned i = 0; i < N_OUT / output_t::size; ++i) {
    output_t ctype = out_local.read();
    for(unsigned j = 0; j < output_t::size; j++) {
      out[i * output_t::size + j] = output_axi_t(ctype[j]);
    }
  }
}
\end{lstlisting}
    \end{minipage}%
    \\
    % \begin{minipage}{0.5\textwidth}
        \centering
        \includegraphics[width=0.5\columnwidth]{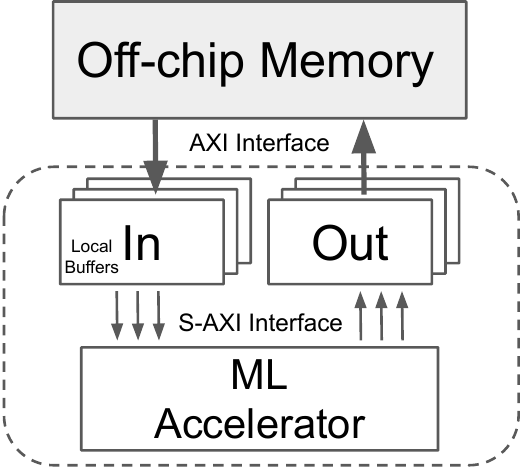}\\
        % \vspace{5px}
        % \includegraphics[width=0.8\columnwidth]{Figs/top_module_axi.png}\\
    % \end{minipage}
\caption{Code for the hls4ml top-level module and accelerator architecture.
\vspace{-0.5cm}}
\label{fig:hls4ml_accelerator}
\end{figure}

The hls4ml flow generates the C++ code listed on the top of Fig.~\ref{fig:hls4ml_accelerator} as the top-level module for the HLS-synthesizable accelerator. 
The module has memory-mapped registers and interfaces (lines 1--4) that allow the accelerator to be programmed, and to read and write data from the off-chip memory, which is shared between the accelerator implemented on the programmable logic and the application running on the processor core. 
The \texttt{bundle} keyword on the \texttt{INTERFACE} pragma specifies the name of the ports as shown in the generated Vivado IP core.
The port \texttt{CTRL\_BUS} groups control registers to program, start, and check the status of the accelerator (line 2).
The bandwidth and throughput of the accelerator can be increased by creating multiple ports (lines 3--4) to load and store data in the local buffers of the accelerators. 
This accelerator style, defined as loose out-of-core with direct memory access to main memory, 
is typical for high-throughput applications that have clear memory access patterns and have input sizes large enough to make vector-processing impractical~\cite{cota2015analysis}.

To further increase performance, the hls4ml accelerator uses a dataflow implementation to produce and consume data in a streaming fashion, thus the local buffers are implemented as FIFOs (line 6--9).
Finally, the interface uses specialized large-width data types, such as the arbitrary precision integer data type \texttt{ap\_int<W>} in Vivado HLS, where \texttt{W} is a data width (64, 128, 256, etc.) to increase the bandwidth of the data communication between the off-chip memory and local buffers. 
The data movers in the HLS code are in charge of unpacking and packing the data to and from the ML processing core of the accelerator (lines 11--19 and 23--28).

While FINN produces an accelerator, which looks similar to the one from hls4ml at a conceptual level, the build process is significantly different.
Both FINN and hls4ml produce a dataflow style accelerator that can be easily integrated into existing designs using the Vivado IP integrator. 
Both accelerators exploit a streaming architecture, which keeps the activation data of the neural network on-chip, thus reducing overall data movement. 
However, while hls4ml builds the whole accelerator from one top level module, FINN builds a similar design by interconnecting multiple IP blocks in Vivado.
Here, each IP block represents one layer of the neural network. 
This means in particular that only the final IP block stitching must be run in series, while the synthesis of each IP block can be done in parallel.

\subsubsection{IP Integration}

We used the Xilinx Vivado Design Suite 2019.1 to instantiate and interconnect IP cores from the Vivado IP catalog and the hls4ml and FINN codesign workflows.
We first interactively used the IP integrator design canvas to develop an automated flow using the Tcl programming interface.
In particular, we integrated most of the IP cores at the advanced extensible interface (AXI) level, but we also worked at the port-and-constraint level to interface the device under test (DUT) with the Embedded Microprocessor Benchmark Consortium (EEMBC) performance and power analysis setup. 

\begin{figure}[tbh!]
    \centering
    \includegraphics[width=0.98\columnwidth]{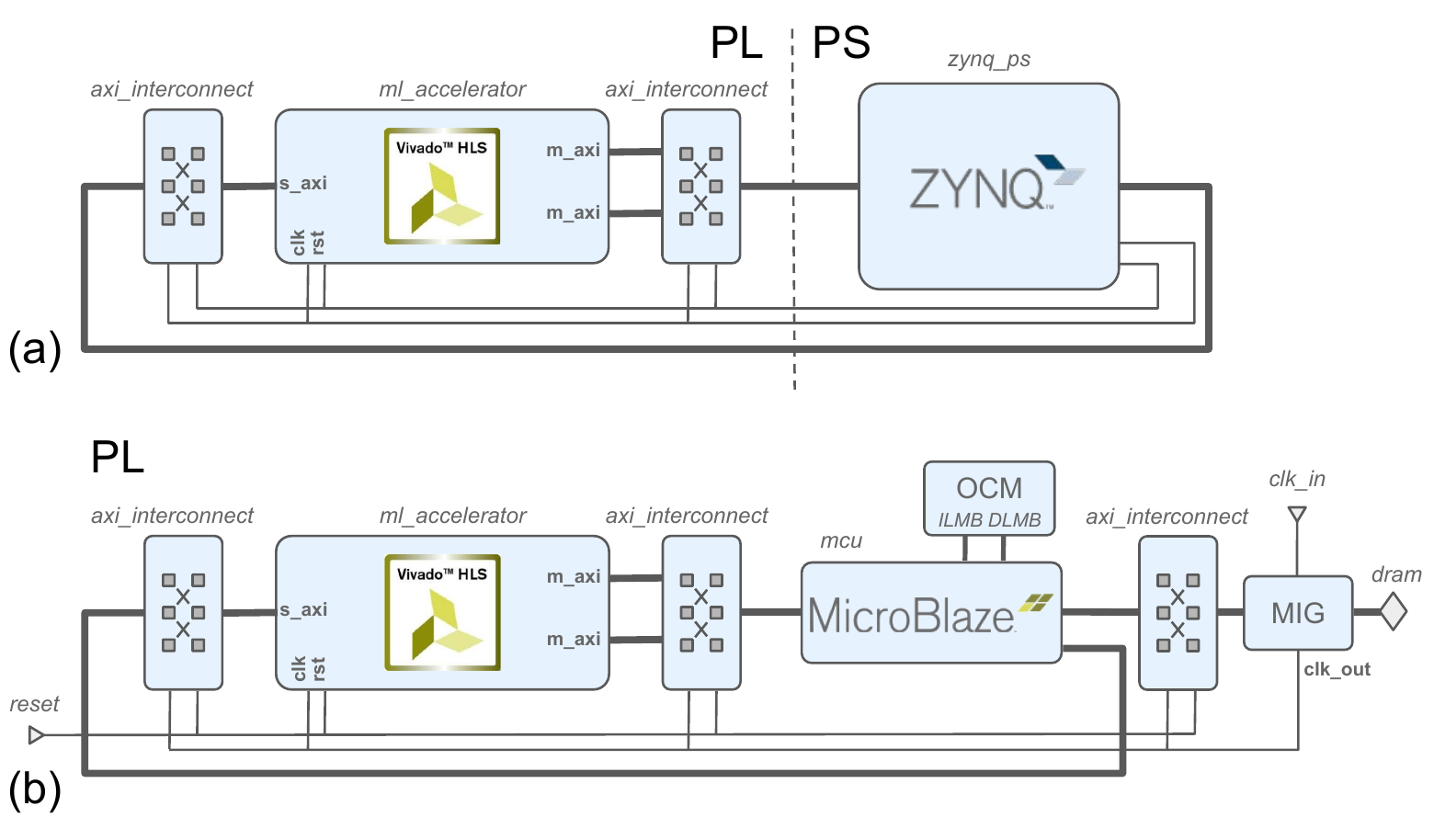}\\
    \caption{Acceleration integration for (a) SoC and (b) FPGA-only designs}
    \label{fig:ip_integration}
\end{figure}

Fig.~\ref{fig:ip_integration} shows the main components for the integration on both Zynq-7020 SoCs and pure FPGA chips. 
A Zynq-7020 SoC combines hard cores, e.g., ARM Cortex-A, of the processing system (PS) and with the flexibility of the programmable logic (PL). 
AXI ports connect the PL with the off-chip memory through the PS.
In Fig.~\ref{fig:ip_integration}a, we integrate both FINN and hls4ml accelerators with AXI buses to support both the accelerator control (\texttt{s\_axi}) and data movement (\texttt{m\_axi}).
Fig.~\ref{fig:ip_integration}b shows the design on a pure FPGA, where we similarly instantiate the accelerator, but we integrate a soft processor (MicroBlaze) on the PL instead. 
The memory controller (MIG) and the on-chip memory (OCM) are instantiated as soft IPs as well. 
For our experiments, we sized the MicroBlaze instruction and data cache in the range 1--16\,kB and the on-chip memory in the range 32--128\,kB to balance BRAM usage and software performance.

\subsubsection{Experimental Results on Development Boards}
In our experimental setup we used two off-the-shelf development boards: the TUL Pynq-Z2 and Digilent Arty A7-100T boards.
%Here we provide details relevant to the our experiments together with final resource usage of the programmable logic.
The TUL Pynq-Z2 board is based on a Xilinx Zynq-7020 SoC and designed for the Xilinx University Program to support the Pynq software stack. 
The Zynq-7020 SoC on the board (\texttt{xc7z020-1clg400c}) combines an ARM dual-core Cortex-A9 processor at 650\,MHz with 13,300 logic slices (four 6-input LUTs and eight FFs), 630\,kB of BRAM, and 220 DSP slices. 
%In addition, the chip is programmable from JTAG, Quad-SPI flash, and MicroSD card. 
%Furthermore, the board is equipped with 512\,MB DDR3 (off-chip memory), 16\,MB Quad-SPI Flash, and various expansion connectors, including two standard Pmod ports.
%Finally, the board is powered by USB or 7--15V external power source.

The Digilent Arty A7-100T board is based on Xilinx Artix-7 technology and designed for low-power and low-cost applications. 
The FPGA chip (\texttt{xc7a100t‐1csg324}) comes with 15,850 logic slices (four 6-input LUTs and eight FFs), 607.5\,kB of BRAM, and 240 DSP slices.
%In addition, the board is equipped with 256\,MB DDR3 for off-chip memory, while the rest of the specifications relevant for the present work are the same as the ones reported for the TUL Pynq-Z2.
Table~\ref{tab:pynqz2_artya7} reports the final resource usage after placement and routing for all designs implemented on both platforms.

We can directly compare the two different solutions---one based on hls4ml and one based on FINN---that were submitted on the same hardware platforms for the IC benchmark task.
First, we note some differences in the model design. 
The hls4ml IC model is a relatively small CNN (58\,115 parameters) implemented using fixed-point precision weights and activations with bit widths in the range 8--12, while the FINN IC model is significantly larger (1\,542\,848 parameters), but implemented with binary weights and activations.
Thus while the FINN IC model implements more operations, they are each less computationally expensive.

Another distinction between the models is the chosen resource-latency tradeoff.
The hls4ml IC model utilizes 58\% fewer BRAMs compared to the FINN IC model for the Pynq-Z2 platform.
However the latency is 18.2 times larger, the bulk of which is required by the penultimate convolutional layer (6.6 times longer latency than the next slowest layer).
The hls4ml streaming architecture chosen is such that the $32\times 32$ input image size is iterated over sequentially.
For each iteration, the inputs are assembled into the corresponding $4\times 4\times 32$ input tensor for a single kernel multiplication and up to 16\,384 multiplications are performed sequentially, resulting in 32 outputs per kernel multiplication.
Thus while the resource usage is kept to a minimum, the worst-case latency scales approximately as $32\times 32 \times 16\,384$ clock cycles.
In future submissions, we plan to more efficiently pipeline these operations to substantially reduce the latency of the hls4ml IC model.

\begin{table*}[th!]
%\footnotesize
  \centering
    \resizebox{\textwidth}{!}{  
    \begin{tabular}{l|rr|rr|rr|rr|rr|r|r} %
      Model & \multicolumn{2}{c|}{LUT} & \multicolumn{2}{c|}{LUTRAM} & \multicolumn{2}{c|}{FF} & \multicolumn{2}{c|}{BRAM [36\,kb]} & \multicolumn{2}{c|}{DSP} & \multicolumn{1}{c|}{Latency [ms]} & \multicolumn{1}{c}{Energy/inf. [$\mu$J]} \\\hline
      \multicolumn{13}{c}{Pynq-Z2} \\\hline
      Available & \multicolumn{2}{c|}{53\,200} & \multicolumn{2}{c|}{17\,400} & \multicolumn{2}{c|}{106\,400} & \multicolumn{2}{c|}{140} & \multicolumn{2}{c|}{220} & -- & -- \\
      \hline
      IC (hls4ml) & 28\,544 & 53.7\% & 3\,756 & 21.6\% & 49\,215 & 46.3\% & 42 & 30.0\% & 4 & 1.8\% & 27.3 & 44\,330 \\
      IC (FINN) &  24\,502 & 46.1\% & 2\,086 & 12.0\% & 34\,354 & 32.3\%& 100 & 71.4\%& 0 & 0.0\% & 1.5 & 2\,535 \\
      AD & 40\,658 & 76.4\% & 3\,659 & 21.0\% & 51\,879 & 48.8\% & 14.5 & 10.4\% & 205 & 93.2\% & 0.019 & 30.1 \\
      KWS & 33\,732 & 63.4\% & 1\,033 & 5.9\% & 34\,405 & 32.3\% & 37 & 26.4\% & 1 & 0.5\% & 0.017 & 30.9 \\ \hline
      \multicolumn{13}{c}{Arty A7-100T} \\\hline
      Available & \multicolumn{2}{c|}{63400} & \multicolumn{2}{c|}{19\,000} & \multicolumn{2}{c|}{126\,800} & \multicolumn{2}{c|}{135} & \multicolumn{2}{c|}{240} & - & - \\
      \hline
      IC (hls4ml) & 39\,126 & 61.7\% & 5\,877 & 30.9\% & 59\,184 & 46.7\% & 50 & 37.0\% & 6 & 2.5\% & 33.1 & 73\,166 \\
      IC (FINN) & 32\,096 & 50.6\% & 3\,154 & 16.6\% & 39\,962 & 31.5\% & 113.5& 84.1\% & 2 & 0.8\% & 1.5 & 3\,419 \\
      AD & 51\,429 & 81.1\% & 5\,780 & 30.4\% & 61\,639 & 48.6\% & 22.5 & 16.7\% & 207 & 86.3\% & 0.045 & 98.4 \\
      KWS & 42\,518 & 67.1\% & 1\,634 & 8.6\% & 43\,157 & 34.0\% & 59.5 & 44.1\% & 2 & 0.8\% & 0.033 & 53.7 \\
    \end{tabular}
    }
    \caption{Resource usage, latency, and energy per inference for the submitted models implemented on the Pynq-Z2 and Arty A7-100T platforms.}
    \label{tab:pynqz2_artya7}
\end{table*}

\subsection{Software Integration}

\subsubsection{Bare-Metal Setup}
%(Giuseppe)}
%\textcolor{red}{Describe the baremetal application, the backend addition to hls4ml, the data compilation, etc. We do not use an RTOS for the specific work but we can leverage it for multi-tasking (multiple accelerators sw-controlled, concurrent pre-processing on CPU/MCU, long batches).}

%The code in Listing~\ref{list:baremetal} shows our bare-metal application.
In our setup, the processor, or microcontroller, is in charge of initiating the memory with the benchmark data, programming the accelerator, starting it, and waiting for its completion with polling on a register. 
Finally, we compare the correctness of the accelerator outputs against precomputed reference outputs.
The accelerator responds to the initial configuration from the processor, and then autonomously transfers data between off-chip memory and local buffers.
The communication between processor and accelerator always uses memory-mapped I/O. 
The processor can directly read and write registers of the accelerator interface that are accessible using pointers in C/C++.

%\begin{lstlisting}[float=htpb,caption="Bare-metal code to control the accelerator.",label={list:baremetal}]
%#define N_SAMPLES /*...*/
%#define N_IN_FEATURES /*...*/
%#define N_OUT_FEATURES /*...*/
%
%/* Data buffers in main memory */
%data_in_t x_data[N_IN_FEATURES * N_SAMPLES] = { /*...*/ };
%data_out_t y_data[N_OUT_FEATURES * N_SAMPLES];
%
%/* Iterate over the samples */
%for (unsigned i = 0; i < N_SAMPLES; i++) {
%    /* Get memory offsets for input and output features */
%    x_data_t* in_features = x_data + i * N_IN_FEATURES;
%    y_data_t* out_features = y_data + i * N_OUT_FEATURES;
%
%    /* Configure the accelerator */
%    XTop_module_axi_Set_in(&accelerator, in_features);
%    XTop_module_axi_Set_out(&accelerator, out_features);
%
%    XTop_module_axi_Start(&accelerator);
%
%    /* Polling */
%    while (!XTop_module_axi_IsDone(&accelerator));
%
%    /* Get error status */
%    hw_flags = XTop_module_axi_Get_return(&accelerator);    
%}
%
%/* Validation */
%/* Compare accelerator outputs with reference values */
%\end{lstlisting}

The EEMBC benchmarking framework requires a serial console connected to the board to view and process the standard output from \texttt{printf} statements. 
We also use a programmable baud rate for both the Zynq and Microblaze designs with an AXI universal asynchronous receiver transmitter (UART) Lite IP for the latter.

\subsection{EEMBC EnergyRunner{\texttrademark} and Test Harness}

\begin{figure}[tbh!]
    \centering
    \includegraphics[width=0.8\columnwidth]{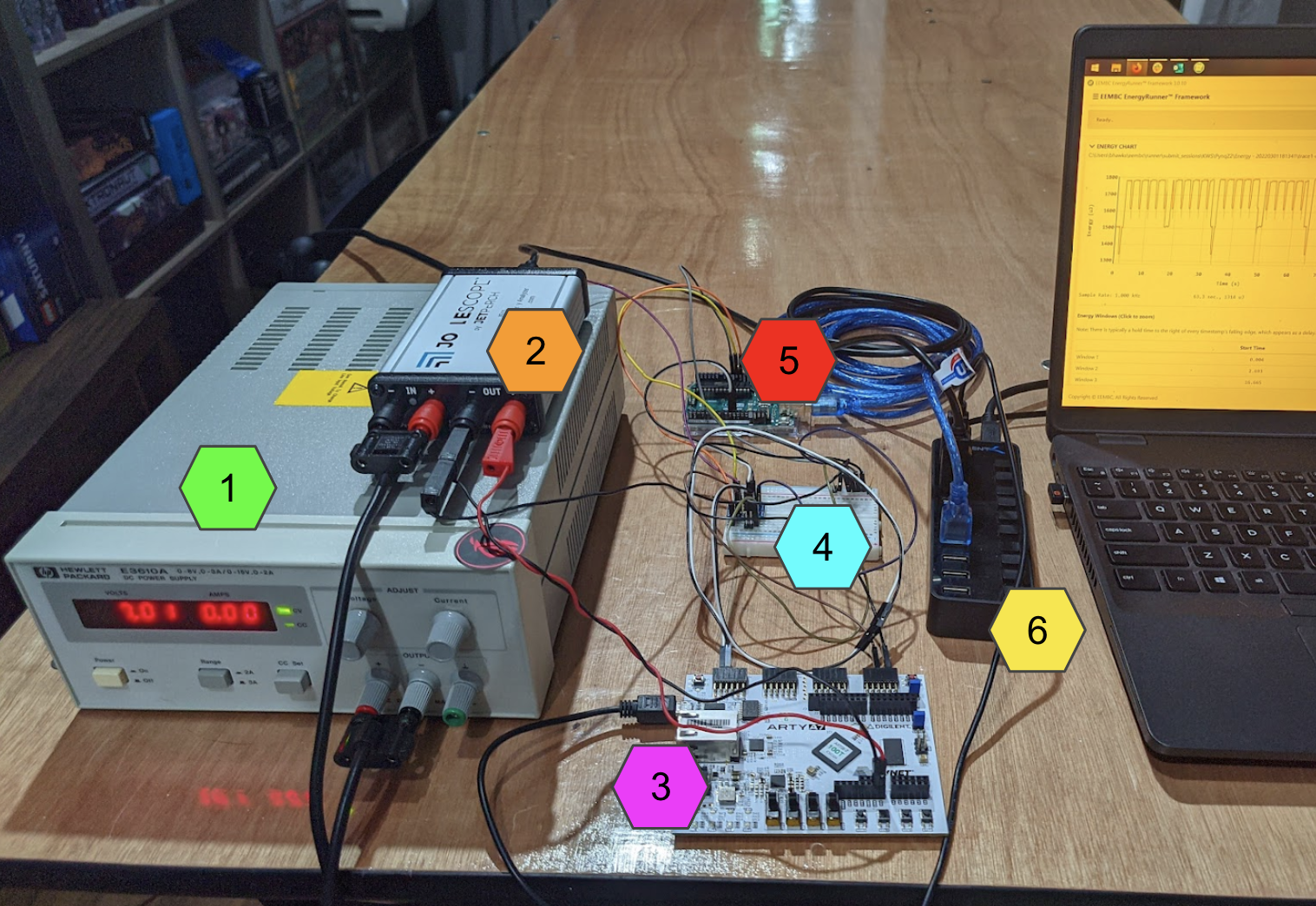}\\
    \vspace{0.5cm}
    \includegraphics[width=0.8\columnwidth]{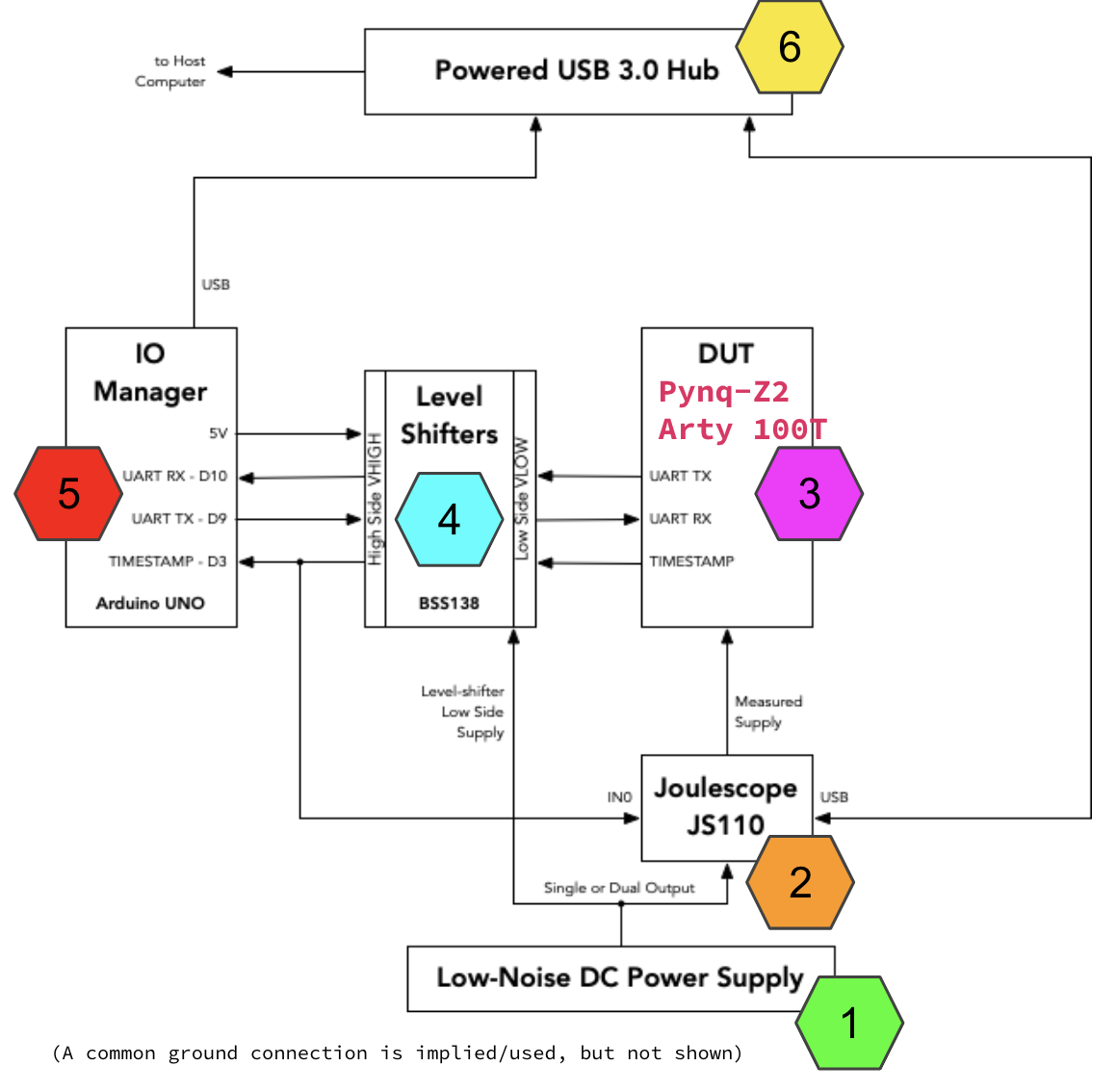}
    \caption{Top: Image of device testing setu. Bottom: Block diagram of testing setup with numerical labels corresponding to the top image.\vspace{-0.7cm}}
    \label{fig:setup}
\end{figure}

In order to submit official results to the benchmark, we must use the benchmarking framework that consists of two pieces of software: the EEMBC EnergyRunner{\texttrademark}~(runner) and test harness. 
The former runs on a host computer and the latter on the DUT. 

The runner application runs on a host computer and communicates over a serial connection with the DUT and other hardware required to perform various benchmark measurements. 
The runner software is responsible for configuring benchmark hardware, sending input samples to the DUT, and calculating benchmark metrics such as latency, network accuracy/AUC, and energy used per inference based on data the DUT and other hardware report.

The test harness is provided as partially implemented C++ code that must be integrated onto the DUT.
The harness communicates with the runner, with functionality including basic command parsing and benchmark-related operations already implemented (and unable to be modified). 
More DUT-specific operations need to be implemented by the submitter, such as loading input data into the accelerator, running a batch size of 1 inference, timer functionality, and general hardware setup.
We implemented the required functionality for the Pynq-Z2 and Arty A7-100T in separate instances of the test harness, and ran the test harness as a bare-metal application, programming it and the applicable bootloader into the DUT's memory to launch the application upon a device restart. 

\subsubsection{Performance and Accuracy Measurements}
When running the benchmark in performance mode, we test the latency of the accelerator, and, as our submission is in the open division, we measure the accuracy over the whole test dataset. 
The physical setup of performance mode consists of the DUT, programmed with the test harness, connected over a serial port to a host computer running the runner application.
In our case, both the Arty A7-100T and Pynq-Z2 were connected over a USB serial connection to a host PC running either Microsoft Windows 10 or Linux. 
For the latency test, a total of 5 samples are sent to the DUT one at a time. 
For each sample, the DUT performs sufficient batch-1 inferences to accumulate at least 10\,s of continuous accelerator run time. 
Once complete, the runner calculates the median latency to perform a single inference over the 5 different samples.
To perform the accuracy test, each sample in the entire test dataset for a given benchmark is sent to the DUT one at a time.
The DUT runs all single-sample inferences, after which the results are returned to the runner to compute the overall accuracy/AUC. 
Logs of both benchmarks are recorded, then submitted along with the code into the official MLPerf Tiny v0.7 GitHub repository.

Table~\ref{tab:pynqz2_artya7} shows the latencies and Table~\ref{tab:summary} shows the measured accuracies.
The submitted designs are comparable in accuracy or AUC to the reference models and have latencies ranging from 30\,ms to 20\,$\mu$s, demonstrating the high throughput achievable with this approach.

\subsubsection{Energy Consumption Measurements}   
To run the benchmark in energy mode, the hardware setup is more complex than in performance mode, as can be seen in Fig.~\ref{fig:setup}. 
It is comprised of a host computer, the DUT, an energy monitor that measures and records the power used over a set interval, an IO Manager (Arduino UNO) running firmware to act as a serial bridge between the host computer and DUT for power isolation and serial port stability when power cycling the DUT, and level shifters between the IO Manager and the DUT. 

Minor modifications are also made to the test harness when running in energy mode. 
First, the baud rate of the DUT is changed from 115\,200 to 9\,600, which is required to communicate with the IO Manager.
Additionally, the time measurement protocol is changed from the DUT's internal timer to holding a GPIO pin that is connected to the energy monitor low for at least 10\,$\mu$s, as the energy monitor manages the timer in this mode.

When running the energy benchmark, the methodology is nearly the same as the latency test, except that the power utilization of the DUT is also recorded by the energy monitor, and the energy per inference is also taken as the median over all of the samples. 
We used the Joulescope JS110 as our energy monitor, and an HP/Agilent E3610A power supply to power the DUT and energy monitor.

The measured energies per inference for each design are shown in Table~\ref{tab:pynqz2_artya7}. 
They vary from 70\,mJ to 30\,$\mu$J per inference depending on the task and hardware platform, demonstrating the relatively low energy consumption possible with our workflows.

% \subsection{Power Optimization (Giuseppe, Ben)}

\section{Summary}
\label{sec:summary}

This paper details the hls4ml-FINN solutions implementing field-programmable gate array (FPGA)-based acceleration for the MLPerf Tiny Inference Benchmark.  
The goal of the submissions was to demonstrate efficient and low-latency solutions on FPGAs using open-source workflows developed by the hls4ml and FINN teams; the solutions also catalyzed collaboration and the development of more accessible tools towards the democratization of powerful tiny ML.

Solutions were provided for the anomaly detection, keyword spotting, and image classification benchmarks.  
Model and design space exploration is presented including performance and hardware implementation optimizations. 
This enabled novel capabilities merged into the tool flows, including layer fusion, FIFO buffer optimization, and the development of the QONNX interchange format for representing flexibly quantized neural networks.  

From this work, we demonstrated a common methodology for building FPGA-optimized ML model implementations targeting different benchmarks.
First, a 32-bit floating-point precision model was trained to determine the baseline expected performance.
Then we generalized the model by introducing additional hyperparameters related to layer sizes, number of layers, pooling choices, and more, and performed a hyperparameter optimization to determine the Pareto-optimal front balancing the performance and inference cost of a given ML model.
Next, we quantized the model from 32-bit floating-point to integer/fixed-point precision reducing the bit width until the model performance began to degrade---the smallest bit width that retained the original baseline performance was then chosen.
Finally, we synthesized the model for the FPGA platforms with further hardware-specific optimizations to fit it within the FPGA resources with minimal latency and maximal throughput.
% These steps were iterated until we found an optimal solution for our hardware platforms.
% Our experience performing this full codesign process using open-source tools is presented.
This approach can be further refined and formalized by integrating with all-in-one, end-to-end workflows like Sherlock~\cite{10.1145/3511472}.

The system-level integration of the AI algorithms is also presented including interfaces to the control platform and setup for measuring performance, latency, and power in the standardized benchmark workflow.  
The optimized designs are comparable in performance to the reference models and have latencies as low as 20\,$\mu$s and measure as low as 30\,$\mu$J per inference. 
The end-to-end workflows are publicly available online\footnote{\url{https://github.com/mlcommons/tiny_results_v0.7/tree/main/open/hls4ml-finn}}.   

\section*{Acknowledgments}
This work was supported by the DOE (Contract No. DE-AC02-07CH11359, Award Nos. DE-SC0021187, DE-SC0021396), the NSF (Cooperative Agreement OAC-2117997, Award No. 1764000), DARPA (C\#: FA8650-18-2-7862), and the ERC (Grant No. 966696).

\bibliographystyle{mlsys2022}
\bibliography{references}

\begin{thebibliography}{61}
\providecommand{\natexlab}[1]{#1}
\providecommand{\url}[1]{\texttt{#1}}
\expandafter\ifx\csname urlstyle\endcsname\relax
  \providecommand{\doi}[1]{doi: #1}\else
  \providecommand{\doi}{doi: \begingroup \urlstyle{rm}\Url}\fi

\bibitem[Aarrestad et~al.(2021)Aarrestad, Loncar, Pierini, Summers, Ngadiuba,
  Petersson, Linander, Iiyama, Guglielmo, Duarte, Harris, Rankin, Jindariani,
  Pedro, Tran, Liu, Kreinar, Wu, and Hoang]{Aarrestad:2021zos}
Aarrestad, T., Loncar, V., Pierini, M., Summers, S., Ngadiuba, J., Petersson,
  C., Linander, H., Iiyama, Y., Guglielmo, G.~D., Duarte, J., Harris, P.,
  Rankin, D., Jindariani, S., Pedro, K., Tran, N., Liu, M., Kreinar, E., Wu,
  Z., and Hoang, D.
\newblock {Fast convolutional neural networks on FPGAs with \texttt{hls4ml}}.
\newblock 2021,
  \href{http://www.arXiv.org/abs/2101.05108}{\texttt{arXiv:2101.05108}}.
\newblock Submitted to \emph{Mach. Learn.: Sci. Technol.}

\bibitem[Abdelouahab et~al.(2018)Abdelouahab, Pelcat, Serot, and
  Berry]{abdelouahab2018accelerating}
Abdelouahab, K., Pelcat, M., Serot, J., and Berry, F.
\newblock Accelerating cnn inference on {FPGAs}: A survey.
\newblock 2018,
  \href{http://www.arXiv.org/abs/1806.01683}{\texttt{arXiv:1806.01683}}.

\bibitem[Bai et~al.(2019)Bai, Lu, Zhang, et~al.]{onnx}
Bai, J., Lu, F., Zhang, K., et~al.
\newblock Onnx: Open neural network exchange.
\newblock \url{https://github.com/onnx/onnx}, 2019.

\bibitem[Banbury et~al.(2021)Banbury, Reddi, Torelli, Holleman, Jeffries,
  Kiraly, Montino, Kanter, Ahmed, Pau, Thakker, Torrini, Warden, Cordaro,
  Guglielmo, Duarte, Gibellini, Parekh, Tran, Tran, Wenxu, and
  Xuesong]{banbury2021mlperf}
Banbury, C., Reddi, V.~J., Torelli, P., Holleman, J., Jeffries, N., Kiraly, C.,
  Montino, P., Kanter, D., Ahmed, S., Pau, D., Thakker, U., Torrini, A.,
  Warden, P., Cordaro, J., Guglielmo, G.~D., Duarte, J., Gibellini, S., Parekh,
  V., Tran, H., Tran, N., Wenxu, N., and Xuesong, X.
\newblock {MLPerf Tiny} benchmark.
\newblock In Vanschoren, J. and Yeung, S. (eds.), \emph{Proceedings of the
  Neural Information Processing Systems Track on Datasets and Benchmarks},
  volume~1, 2021,
  \href{http://www.arXiv.org/abs/2106.07597}{\texttt{arXiv:2106.07597}}.
\newblock URL
  \url{https://datasets-benchmarks-proceedings.neurips.cc/paper/2021/file/da4fb5c6e93e74d3df8527599fa62642-Paper-round1.pdf}.

\bibitem[Banbury et~al.(2020)Banbury, Reddi, Lam, Fu, Fazel, Holleman, Huang,
  Hurtado, Kanter, Lokhmotov, Patterson, Pau, Seo, Sieracki, Thakker, Verhelst,
  and Yadav]{tinymlbench}
Banbury, C.~R., Reddi, V.~J., Lam, M., Fu, W., Fazel, A., Holleman, J., Huang,
  X., Hurtado, R., Kanter, D., Lokhmotov, A., Patterson, D., Pau, D., Seo,
  J.-s., Sieracki, J., Thakker, U., Verhelst, M., and Yadav, P.
\newblock Benchmarking tinyml systems: Challenges and direction.
\newblock 2020,
  \href{http://www.arXiv.org/abs/2003.04821}{\texttt{arXiv:2003.04821}}.

\bibitem[Baskin et~al.(2018)Baskin, Schwartz, Zheltonozhskii, Liss, Giryes,
  Bronstein, and Mendelson]{bops}
Baskin, C., Schwartz, E., Zheltonozhskii, E., Liss, N., Giryes, R., Bronstein,
  A.~M., and Mendelson, A.
\newblock {UNIQ}: Uniform noise injection for the quantization of neural
  networks.
\newblock 2018,
  \href{http://www.arXiv.org/abs/1804.10969}{\texttt{arXiv:1804.10969}}.

\bibitem[Blott et~al.(2018{\natexlab{a}})Blott, Preu{\ss}er, Fraser,
  Gambardella, O'Brien, and Umuroglu]{blott2018finnr}
Blott, M., Preu{\ss}er, T., Fraser, N., Gambardella, G., O'Brien, K., and
  Umuroglu, Y.
\newblock {FINN-R}: An end-to-end deep-learning framework for fast exploration
  of quantized neural networks.
\newblock \emph{ACM Trans. Reconfigurable Technol. Syst.}, 11\penalty0 (3),
  2018{\natexlab{a}}.
\newblock ISSN 1936-7406,
  \href{http://doi.org/10.1145/3242897}{\texttt{doi:10.1145/3242897}},
  \href{http://www.arXiv.org/abs/1809.04570}{\texttt{arXiv:1809.04570}}.

\bibitem[Blott et~al.(2018{\natexlab{b}})Blott, Preu{\ss}er, Fraser,
  Gambardella, O'brien, Umuroglu, Leeser, and Vissers]{blott2018finn}
Blott, M., Preu{\ss}er, T.~B., Fraser, N.~J., Gambardella, G., O'brien, K.,
  Umuroglu, Y., Leeser, M., and Vissers, K.
\newblock Finn-r: An end-to-end deep-learning framework for fast exploration of
  quantized neural networks.
\newblock \emph{ACM Transactions on Reconfigurable Technology and Systems
  (TRETS)}, 11\penalty0 (3):\penalty0 1, 2018{\natexlab{b}}.

\bibitem[Chang et~al.(2021)Chang, Li, Sun, Shi, So, Qian, Wang, and
  Lin]{chang2020mix}
Chang, S.-E., Li, Y., Sun, M., Shi, R., So, H. K.~H., Qian, X., Wang, Y., and
  Lin, X.
\newblock Mix and match: A novel {FPGA}-centric deep neural network
  quantization framework.
\newblock In \emph{27th IEEE International Symposium on High-Performance
  Computer Architecture (HPCA)}, 2021,
  \href{http://www.arXiv.org/abs/2012.04240}{\texttt{arXiv:2012.04240}}.

\bibitem[Coelho et~al.(2021)Coelho, Kuusela, Li, Zhuang, Aarrestad, Loncar,
  Ngadiuba, Pierini, Pol, and Summers]{Coelho:2020zfu}
Coelho, C.~N., Kuusela, A., Li, S., Zhuang, H., Aarrestad, T., Loncar, V.,
  Ngadiuba, J., Pierini, M., Pol, A.~A., and Summers, S.
\newblock {Automatic deep heterogeneous quantization of deep neural networks
  for ultra low-area, low-latency inference on the edge at particle colliders}.
\newblock \emph{Nat. Mach. Intell.}, 3:\penalty0 675, 2021,
  \href{http://www.arXiv.org/abs/2006.10159}{\texttt{arXiv:2006.10159}}.

\bibitem[Cota et~al.(2015)Cota, Mantovani, Di~Guglielmo, and
  Carloni]{cota2015analysis}
Cota, E.~G., Mantovani, P., Di~Guglielmo, G., and Carloni, L.~P.
\newblock An analysis of accelerator coupling in heterogeneous architectures.
\newblock In \emph{ACM/EDAC/IEEE Design Automation Conference (DAC)}, pp.\ ~1.
  IEEE, 2015.

\bibitem[Deiana et~al.(2022)Deiana, Tran, et~al.]{Deiana:2021niw}
Deiana, A.~M., Tran, N., et~al.
\newblock {Applications and Techniques for Fast Machine Learning in Science}.
\newblock \emph{Front. Big Data}, 5:\penalty0 787421, 2022,
  \href{http://doi.org/10.3389/fdata.2022.787421}{\texttt{doi:10.3389/fdata.2022.787421}},
  \href{http://www.arXiv.org/abs/2110.13041}{\texttt{arXiv:2110.13041}}.

\bibitem[{Determined AI}(2018)]{det-ai}
{Determined AI}.
\newblock Determined ai, 2018.
\newblock URL \url{https://www.determined.ai}.

\bibitem[{DiCecco} et~al.(2016){DiCecco}, {Lacey}, {Vasiljevic}, {Chow},
  {Taylor}, and {Areibi}]{caffeinatedFPGAs}
{DiCecco}, R., {Lacey}, G., {Vasiljevic}, J., {Chow}, P., {Taylor}, G., and
  {Areibi}, S.
\newblock Caffeinated fpgas: Fpga framework for convolutional neural networks.
\newblock In \emph{2016 International Conference on Field-Programmable
  Technology (FPT)}, pp.\  265. IEEE, 2016,
  \href{http://doi.org/10.1109/FPT.2016.7929549}{\texttt{doi:10.1109/FPT.2016.7929549}},
  \href{http://www.arXiv.org/abs/1609.09671}{\texttt{arXiv:1609.09671}}.

\bibitem[Duarte et~al.(2018)Duarte, Han, et~al.]{Duarte:2018ite}
Duarte, J., Han, S., et~al.
\newblock {Fast inference of deep neural networks in FPGAs for particle
  physics}.
\newblock \emph{JINST}, 13:\penalty0 P07027, 2018,
  \href{http://doi.org/10.1088/1748-0221/13/07/P07027}{\texttt{doi:10.1088/1748-0221/13/07/P07027}},
  \href{http://www.arXiv.org/abs/1804.06913}{\texttt{arXiv:1804.06913}}.

\bibitem[Elabd et~al.(2022)]{Elabd:2021lgo}
Elabd, A. et~al.
\newblock {Graph Neural Networks for Charged Particle Tracking on FPGAs}.
\newblock \emph{Front. Big Data}, 5:\penalty0 828666, 2022,
  \href{http://doi.org/10.3389/fdata.2022.828666}{\texttt{doi:10.3389/fdata.2022.828666}},
  \href{http://www.arXiv.org/abs/2112.02048}{\texttt{arXiv:2112.02048}}.

\bibitem[Fahim et~al.(2021)Fahim, Hawks, et~al.]{Fahim:2021cic}
Fahim, F., Hawks, B., et~al.
\newblock {hls4ml: An Open-Source Codesign Workflow to Empower Scientific
  Low-Power Machine Learning Devices}.
\newblock In \emph{{tinyML Research Symposium 2021}}, 3 2021,
  \href{http://www.arXiv.org/abs/2103.05579}{\texttt{arXiv:2103.05579}}.

\bibitem[Gautier et~al.(2022)Gautier, Althoff, Crutchfield, and
  Kastner]{10.1145/3511472}
Gautier, Q., Althoff, A., Crutchfield, C.~L., and Kastner, R.
\newblock Sherlock: A multi-objective design space exploration framework.
\newblock \emph{ACM Trans. Des. Autom. Electron. Syst.}, 27\penalty0 (4), 2022,
  \href{http://doi.org/10.1145/3511472}{\texttt{doi:10.1145/3511472}}.

\bibitem[{Gokhale} et~al.(2017){Gokhale}, {Zaidy}, {Chang}, and
  {Culurciello}]{snowflake}
{Gokhale}, V., {Zaidy}, A., {Chang}, A. X.~M., and {Culurciello}, E.
\newblock Snowflake: An efficient hardware accelerator for convolutional neural
  networks.
\newblock In \emph{2017 IEEE International Symposium on Circuits and Systems
  (ISCAS)}, pp.\ ~1, 2017,
  \href{http://doi.org/10.1109/ISCAS.2017.8050809}{\texttt{doi:10.1109/ISCAS.2017.8050809}},
  \href{http://www.arXiv.org/abs/1708.02579}{\texttt{arXiv:1708.02579}}.

\bibitem[{Google}(2020)]{qkeras}
{Google}.
\newblock Qkeras, 2020.
\newblock URL \url{https://github.com/google/qkeras}.

\bibitem[Guan et~al.(2017)Guan, Liang, Xu, Wang, Shi, Chen, Sun, Zhang, and
  Cong]{fpdnn}
Guan, Y., Liang, H., Xu, N., Wang, W., Shi, S., Chen, X., Sun, G., Zhang, W.,
  and Cong, J.
\newblock {FP-DNN}: An automated framework for mapping deep neural networks
  onto {FPGAs} with {RTL-HLS} hybrid templates.
\newblock In \emph{2017 IEEE 25th Annual International Symposium on
  Field-Programmable Custom Computing Machines (FCCM)}, pp.\  152, 2017,
  \href{http://doi.org/10.1109/FCCM.2017.25}{\texttt{doi:10.1109/FCCM.2017.25}}.

\bibitem[Guo et~al.(2019)Guo, Zeng, Yu, Wang, and Yang]{10.1145/3289185}
Guo, K., Zeng, S., Yu, J., Wang, Y., and Yang, H.
\newblock A survey of {FPGA}-based neural network inference accelerators.
\newblock \emph{ACM Trans. Reconfigurable Technol. Syst.}, 12, 2019.
\newblock ISSN 1936-7406,
  \href{http://doi.org/10.1145/3289185}{\texttt{doi:10.1145/3289185}},
  \href{http://www.arXiv.org/abs/1712.08934}{\texttt{arXiv:1712.08934}}.

\bibitem[Hacene et~al.(2020)Hacene, Gripon, Arzel, Farrugia, and
  Bengio]{hacene2018quantized}
Hacene, G.~B., Gripon, V., Arzel, M., Farrugia, N., and Bengio, Y.
\newblock Quantized guided pruning for efficient hardware implementations of
  convolutional neural networks.
\newblock In \emph{2020 18th IEEE International New Circuits and Systems
  Conference (NEWCAS)}, pp.\  206, 2020,
  \href{http://doi.org/10.1109/NEWCAS49341.2020.9159769}{\texttt{doi:10.1109/NEWCAS49341.2020.9159769}},
  \href{http://www.arXiv.org/abs/1812.11337}{\texttt{arXiv:1812.11337}}.

\bibitem[Hawks et~al.(2021)Hawks, Duarte, Fraser, Pappalardo, Tran, and
  Umuroglu]{Hawks:2021ruw}
Hawks, B., Duarte, J., Fraser, N.~J., Pappalardo, A., Tran, N., and Umuroglu,
  Y.
\newblock {Ps and Qs: Quantization-aware pruning for efficient low latency
  neural network inference}.
\newblock \emph{Front. AI}, 4:\penalty0 676564, 2021,
  \href{http://doi.org/10.3389/frai.2021.676564}{\texttt{doi:10.3389/frai.2021.676564}},
  \href{http://www.arXiv.org/abs/2102.11289}{\texttt{arXiv:2102.11289}}.

\bibitem[He et~al.(2016)He, Zhang, Ren, and Sun]{resnet}
He, K., Zhang, X., Ren, S., and Sun, J.
\newblock Deep residual learning for image recognition.
\newblock In \emph{2016 IEEE Conference on Computer Vision and Pattern
  Recognition (CVPR)}, pp.\  770, 2016,
  \href{http://doi.org/10.1109/CVPR.2016.90}{\texttt{doi:10.1109/CVPR.2016.90}},
  \href{http://www.arXiv.org/abs/1512.03385}{\texttt{arXiv:1512.03385}}.

\bibitem[Hubara et~al.(2016)Hubara, Courbariaux, Soudry, El-Yaniv, and
  Bengio]{binaryNet}
Hubara, I., Courbariaux, M., Soudry, D., El-Yaniv, R., and Bengio, Y.
\newblock Binarized neural networks.
\newblock In Lee, D., Sugiyama, M., Luxburg, U., Guyon, I., and Garnett, R.
  (eds.), \emph{Advances in Neural Information Processing Systems}, volume~29.
  Curran Associates, Inc., 2016,
  \href{http://www.arXiv.org/abs/1602.02830}{\texttt{arXiv:1602.02830}}.
\newblock URL
  \url{https://proceedings.neurips.cc/paper/2016/file/d8330f857a17c53d217014ee776bfd50-Paper.pdf}.

\bibitem[Iiyama et~al.(2021)]{Iiyama:2020wap}
Iiyama, Y. et~al.
\newblock Distance-weighted graph neural networks on {FPGAs} for real-time
  particle reconstruction in high energy physics.
\newblock \emph{Front. Big Data}, 3:\penalty0 598927, 2021,
  \href{http://doi.org/10.3389/fdata.2020.598927}{\texttt{doi:10.3389/fdata.2020.598927}},
  \href{http://www.arXiv.org/abs/2008.03601}{\texttt{arXiv:2008.03601}}.

\bibitem[{International Telecommunication Union}(2021)]{itu}
{International Telecommunication Union}.
\newblock {ITU-ML5G-PS-007: Lightning-Fast Modulation Classification with
  Hardware-Efficient Neural Networks}.
\newblock \url{https://challenge.aiforgood.itu.int/match/matchitem/34}, 2021.
\newblock Accessed: 2022-03-31.

\bibitem[Kagermann et~al.(2013)Kagermann, Wahlster, and
  Helbig]{KagermannWahlsterHelbig2013en}
Kagermann, H., Wahlster, W., and Helbig, J.
\newblock Recommendations for implementing the strategic initiative industrie
  4.0 -- securing the future of german manufacturing industry.
\newblock Final report of the industrie 4.0 working group, acatech -- National
  Academy of Science and Engineering, M\"{u}nchen, 2013.
\newblock URL
  \url{http://forschungsunion.de/pdf/industrie_4_0_final_report.pdf}.

\bibitem[Koizumi et~al.(2019)Koizumi, Saito, Uematsu, Harada, and
  Imoto]{koizumi2019toyadmos}
Koizumi, Y., Saito, S., Uematsu, H., Harada, N., and Imoto, K.
\newblock Toyadmos: A dataset of miniature-machine operating sounds for
  anomalous sound detection.
\newblock In \emph{2019 IEEE Workshop on Applications of Signal Processing to
  Audio and Acoustics (WASPAA)}, pp.\  313. IEEE, 2019.

\bibitem[Krizhevsky et~al.(2009)Krizhevsky, Nair, and Hinton]{cifar10}
Krizhevsky, A., Nair, V., and Hinton, G.
\newblock Cifar-10 (canadian institute for advanced research), 2009.
\newblock URL \url{http://www.cs.toronto.edu/~kriz/cifar.html}.

\bibitem[Li et~al.(2020)Li, Jamieson, Rostamizadeh, Gonina, Ben-tzur, Hardt,
  Recht, and Talwalkar]{aASHA}
Li, L., Jamieson, K., Rostamizadeh, A., Gonina, E., Ben-tzur, J., Hardt, M.,
  Recht, B., and Talwalkar, A.
\newblock A system for massively parallel hyperparameter tuning.
\newblock In Dhillon, I., Papailiopoulos, D., and Sze, V. (eds.),
  \emph{Proceedings of Machine Learning and Systems}, volume~2, pp.\  230,
  2020, \href{http://www.arXiv.org/abs/1810.05934}{\texttt{arXiv:1810.05934}}.
\newblock URL
  \url{https://proceedings.mlsys.org/paper/2020/file/f4b9ec30ad9f68f89b29639786cb62ef-Paper.pdf}.

\bibitem[Majumder \& Bondhugula(2019)Majumder and
  Bondhugula]{majumder2019flexible}
Majumder, K. and Bondhugula, U.
\newblock A flexible {FPGA} accelerator for convolutional neural networks.
\newblock 2019,
  \href{http://www.arXiv.org/abs/1912.07284}{\texttt{arXiv:1912.07284}}.

\bibitem[Muhizi(2021)]{qdensebatchnorm}
Muhizi, J.
\newblock Add support for qdense\_batchnorm in qkeras, 2021.
\newblock URL \url{https://github.com/google/qkeras/pull/74}.

\bibitem[Ngadiuba et~al.(2020)Ngadiuba, Loncar, Pierini, Summers, Di~Guglielmo,
  Duarte, Harris, Rankin, Jindariani, Liu, and et~al.]{DiGuglielmo:2020eqx}
Ngadiuba, J., Loncar, V., Pierini, M., Summers, S., Di~Guglielmo, G., Duarte,
  J., Harris, P., Rankin, D., Jindariani, S., Liu, M., and et~al.
\newblock Compressing deep neural networks on {FPGAs} to binary and ternary
  precision with \texttt{hls4ml}.
\newblock \emph{Mach. Learn.: Sci. Technol.}, 2:\penalty0 015001, 2020,
  \href{http://doi.org/10.1088/2632-2153/aba042}{\texttt{doi:10.1088/2632-2153/aba042}},
  \href{http://www.arXiv.org/abs/2003.06308}{\texttt{arXiv:2003.06308}}.

\bibitem[O'Malley et~al.(2019)O'Malley, Bursztein, Long, Chollet, Jin,
  Invernizzi, et~al.]{omalley2019kerastuner}
O'Malley, T., Bursztein, E., Long, J., Chollet, F., Jin, H., Invernizzi, L.,
  et~al.
\newblock Kerastuner.
\newblock \url{https://github.com/keras-team/keras-tuner}, 2019.

\bibitem[Pappalardo(2021)]{brevitas}
Pappalardo, A.
\newblock Xilinx/brevitas, 2021,
  \href{http://doi.org/10.5281/zenodo.3333552}{\texttt{doi:10.5281/zenodo.3333552}}.
\newblock URL \url{https://github.com/xilinx/brevitas}.

\bibitem[Pappalardo et~al.(2022)Pappalardo, Umuroglu, et~al.]{qonnx_paper}
Pappalardo, A., Umuroglu, Y., et~al.
\newblock {QONNX: Representing Arbitrary-Precision Quantized Neural Networks}.
\newblock In \emph{{4th Workshop on Accelerated Machine Learning (AccML) at
  HiPEAC 2022 Conference}}, 2022,
  \href{http://www.arXiv.org/abs/2206.07527}{\texttt{arXiv:2206.07527}}.
\newblock URL
  \url{https://accml.dcs.gla.ac.uk/papers/2022/4thAccML_paper_1(12).pdf}.

\bibitem[Paszke et~al.(2019)Paszke, Gross, Massa, Lerer, Bradbury, Chanan,
  Killeen, Lin, Gimelshein, Antiga, Desmaison, Kopf, Yang, DeVito, Raison,
  Tejani, Chilamkurthy, Steiner, Fang, Bai, and Chintala]{pytorch}
Paszke, A., Gross, S., Massa, F., Lerer, A., Bradbury, J., Chanan, G., Killeen,
  T., Lin, Z., Gimelshein, N., Antiga, L., Desmaison, A., Kopf, A., Yang, E.,
  DeVito, Z., Raison, M., Tejani, A., Chilamkurthy, S., Steiner, B., Fang, L.,
  Bai, J., and Chintala, S.
\newblock Pytorch: An imperative style, high-performance deep learning library.
\newblock In Wallach, H., Larochelle, H., Beygelzimer, A., d\textquotesingle
  Alch\'{e}-Buc, F., Fox, E., and Garnett, R. (eds.), \emph{Advances in Neural
  Information Processing Systems 32}, pp.\  8024--8035. Curran Associates,
  Inc., 2019.
\newblock URL
  \url{http://papers.neurips.cc/paper/9015-pytorch-an-imperative-style-high-performance-deep-learning-library.pdf}.

\bibitem[{Rahman} et~al.(2016){Rahman}, {Lee}, and {Choi}]{7459526}
{Rahman}, A., {Lee}, J., and {Choi}, K.
\newblock Efficient {FPGA} acceleration of convolutional neural networks using
  logical-{3D} compute array.
\newblock In \emph{2016 Design, Automation Test in Europe Conference Exhibition
  (DATE)}, pp.\  1393. IEEE, 2016.

\bibitem[Sharma et~al.(2016)Sharma, Park, Mahajan, Amaro, Kim, Shao, Mishra,
  and Esmaeilzadeh]{dnnweaver:micro16}
Sharma, H., Park, J., Mahajan, D., Amaro, E., Kim, J.~K., Shao, C., Mishra, A.,
  and Esmaeilzadeh, H.
\newblock From high-level deep neural models to fpgas.
\newblock In \emph{49th Annual IEEE/ACM International Symposium on
  Microarchitecture (MICRO)}, pp.\ ~1. IEEE, 2016.

\bibitem[Shawahna et~al.(2019)Shawahna, Sait, and El-Maleh]{Shawahna_2019}
Shawahna, A., Sait, S.~M., and El-Maleh, A.
\newblock {FPGA}-based accelerators of deep learning networks for learning and
  classification: {A} review.
\newblock \emph{IEEE Access}, 7:\penalty0 7823, 2019.
\newblock ISSN 2169-3536,
  \href{http://doi.org/10.1109/access.2018.2890150}{\texttt{doi:10.1109/access.2018.2890150}},
  \href{http://www.arXiv.org/abs/1901.00121}{\texttt{arXiv:1901.00121}}.

\bibitem[Simonyan \& Zisserman(2015)Simonyan and Zisserman]{VGG}
Simonyan, K. and Zisserman, A.
\newblock Very deep convolutional networks for large-scale image recognition.
\newblock In Bengio, Y. and LeCun, Y. (eds.), \emph{3rd International
  Conference on Learning Representations, {ICLR} 2015, Conference Track
  Proceedings}, 2015,
  \href{http://www.arXiv.org/abs/1409.1556}{\texttt{arXiv:1409.1556}}.

\bibitem[Summers et~al.(2020)]{Summers:2020xiy}
Summers, S. et~al.
\newblock Fast inference of boosted decision trees in {FPGAs} for particle
  physics.
\newblock \emph{JINST}, 15:\penalty0 P05026, 2020,
  \href{http://doi.org/10.1088/1748-0221/15/05/P05026}{\texttt{doi:10.1088/1748-0221/15/05/P05026}},
  \href{http://www.arXiv.org/abs/2002.02534}{\texttt{arXiv:2002.02534}}.

\bibitem[{tinyML Foundation}(2019)]{tiny}
{tinyML Foundation}.
\newblock {About}, 2019.
\newblock URL \url{https://www.tinyml.org}.

\bibitem[Torralba et~al.(2008)Torralba, Fergus, and Freeman]{torralba200880}
Torralba, A., Fergus, R., and Freeman, W.~T.
\newblock 80 million tiny images: A large data set for nonparametric object and
  scene recognition.
\newblock \emph{IEEE transactions on pattern analysis and machine
  intelligence}, 30\penalty0 (11):\penalty0 1958, 2008.

\bibitem[Umuroglu \& Jahre(2017)Umuroglu and Jahre]{streamlining}
Umuroglu, Y. and Jahre, M.
\newblock Streamlined deployment for quantized neural networks.
\newblock In \emph{International Workshop on Highly Efficient Neural Networks
  Design (HENND)}, 2017,
  \href{http://www.arXiv.org/abs/1709.04060}{\texttt{arXiv:1709.04060}}.

\bibitem[Umuroglu et~al.(2017)Umuroglu, Fraser, Gambardella, Blott, Leong,
  Jahre, and Vissers]{finn}
Umuroglu, Y., Fraser, N.~J., Gambardella, G., Blott, M., Leong, P., Jahre, M.,
  and Vissers, K.
\newblock Finn: A framework for fast, scalable binarized neural network
  inference.
\newblock In \emph{Proceedings of the 2017 ACM/SIGDA International Symposium on
  Field-Programmable Gate Arrays}, FPGA '17, pp.\ ~65. ACM, 2017.

\bibitem[Venieris \& Bouganis(2016)Venieris and Bouganis]{venieris2016fccm}
Venieris, S.~I. and Bouganis, C.-S.
\newblock {fpgaConvNet: A framework for mapping convolutional neural networks
  on FPGAs}.
\newblock In \emph{2016 {IEEE} 24th Annual International Symposium on
  Field-Programmable Custom Computing Machines ({FCCM})}, pp.\ ~40. IEEE, 2016,
  \href{http://doi.org/10.1109/FCCM.2016.22}{\texttt{doi:10.1109/FCCM.2016.22}}.

\bibitem[Venieris \& Bouganis(2017{\natexlab{a}})Venieris and
  Bouganis]{venieris2017fpga}
Venieris, S.~I. and Bouganis, C.-S.
\newblock {fpgaConvNet: Automated mapping of convolutional neural networks on
  FPGAs}.
\newblock In \emph{Proceedings of the 2017 ACM/SIGDA International Symposium on
  Field-Programmable Gate Arrays}, pp.\  291. ACM, 2017{\natexlab{a}}.
\newblock ISBN 978-1-4503-4354-1,
  \href{http://doi.org/10.1145/3020078.3021791}{\texttt{doi:10.1145/3020078.3021791}}.

\bibitem[Venieris \& Bouganis(2017{\natexlab{b}})Venieris and
  Bouganis]{venieris2017fpgaconvnet}
Venieris, S.~I. and Bouganis, C.-S.
\newblock {fpgaConvNet}: A toolflow for mapping diverse convolutional neural
  networks on embedded {FPGAs}.
\newblock In \emph{NIPS 2017 Workshop on Machine Learning on the Phone and
  other Consumer Devices}, 2017{\natexlab{b}},
  \href{http://www.arXiv.org/abs/1711.08740}{\texttt{arXiv:1711.08740}}.

\bibitem[Venieris \& Bouganis(2017{\natexlab{c}})Venieris and
  Bouganis]{venieris2017fpl}
Venieris, S.~I. and Bouganis, C.~S.
\newblock {Latency-driven design for FPGA-based convolutional neural networks}.
\newblock In \emph{2017 27th International Conference on Field Programmable
  Logic and Applications (FPL)}, pp.\ ~1, 2017{\natexlab{c}},
  \href{http://doi.org/10.23919/FPL.2017.8056828}{\texttt{doi:10.23919/FPL.2017.8056828}}.

\bibitem[{Venieris} et~al.(2018){Venieris}, {Kouris}, and
  {Bouganis}]{2018arXiv180305900V}
{Venieris}, S.~I., {Kouris}, A., and {Bouganis}, C.-S.
\newblock Toolflows for mapping convolutional neural networks on {FPGAs}: A
  survey and future directions.
\newblock \emph{ACM Comput. Surv.}, 51, 2018.
\newblock ISSN 0360-0300,
  \href{http://doi.org/10.1145/3186332}{\texttt{doi:10.1145/3186332}},
  \href{http://www.arXiv.org/abs/1803.05900}{\texttt{arXiv:1803.05900}}.

\bibitem[{Warden}(2018)]{speechcommandsv2}
{Warden}, P.
\newblock {Speech Commands: A Dataset for Limited-Vocabulary Speech
  Recognition}.
\newblock \emph{ArXiv e-prints}, April 2018,
  \href{http://www.arXiv.org/abs/1804.03209}{\texttt{arXiv:1804.03209}}.
\newblock URL \url{https://arxiv.org/abs/1804.03209}.

\bibitem[Whatmough et~al.(2019{\natexlab{a}})Whatmough, Zhou, Hansen,
  Venkataramanaiah, sun Seo, and Mattina]{whatmough2019fixynn}
Whatmough, P.~N., Zhou, C., Hansen, P., Venkataramanaiah, S.~K., sun Seo, J.,
  and Mattina, M.
\newblock {FixyNN}: Efficient hardware for mobile computer vision via transfer
  learning.
\newblock In \emph{2nd SysML Conference}, 2019{\natexlab{a}},
  \href{http://www.arXiv.org/abs/1902.11128}{\texttt{arXiv:1902.11128}}.
\newblock URL \url{https://mlsys.org/Conferences/2019/doc/2019/69.pdf}.

\bibitem[Whatmough et~al.(2019{\natexlab{b}})]{deepfreeze}
Whatmough, P.~N. et~al.
\newblock Arm-software/deepfreeze.
\newblock \url{https://github.com/ARM-software/DeepFreeze}, 2019{\natexlab{b}}.

\bibitem[Wright et~al.(2020)Wright, Pit-Claude, Miller, and
  @threonorm]{pyverilator}
Wright, A., Pit-Claude, C., Miller, D., and @threonorm.
\newblock Pyverilator, 2020.
\newblock URL \url{https://github.com/csail-csg/pyverilator}.

\bibitem[{Xilinx}(2021)]{qonnx}
{Xilinx}.
\newblock {QONNX and FINN}.
\newblock \url{https://xilinx.github.io/finn//2021/11/03/qonnx-and-finn.html},
  2021.

\bibitem[Xilinx(2021)]{vitisai}
Xilinx.
\newblock Xilinx/vitis-ai.
\newblock \url{https://github.com/Xilinx/Vitis-AI}, 2021.

\bibitem[Yoshioka(2020)]{kerop}
Yoshioka, K.
\newblock kentaroy47/keras-opcounter, 2020.
\newblock URL \url{https://github.com/kentaroy47/keras-Opcounter}.

\bibitem[Zhang et~al.(2017)Zhang, Suda, Lai, and Chandra]{helloEdge}
Zhang, Y., Suda, N., Lai, L., and Chandra, V.
\newblock {Hello Edge: Keyword Spotting on Microcontroller}.
\newblock 2017,
  \href{http://www.arXiv.org/abs/1711.07128}{\texttt{arXiv:1711.07128}}.

\end{thebibliography}

\end{document}